\documentclass[twoside]{article}

\usepackage[preprint]{aistats2024}

\usepackage{url}
\usepackage{graphicx,color}
\usepackage{amsmath,amssymb,amsfonts}
\usepackage{algorithmic}
\usepackage{cite}
\usepackage[hidelinks]{hyperref}
\usepackage[resetlabels]{multibib}
\usepackage{titletoc}       
\usepackage{enumitem}       
\usepackage{dblfloatfix}    
\usepackage{booktabs}
\usepackage{bbding}         

\usepackage{amsthm,amsmath,amsfonts,amssymb}
\usepackage{bbm}

\usepackage{bm}

\def\mP{{\bm{P}}}

\DeclareMathAlphabet{\mathsfit}{\encodingdefault}{\sfdefault}{m}{sl}
\SetMathAlphabet{\mathsfit}{bold}{\encodingdefault}{\sfdefault}{bx}{n}

\newtheorem{theorem}{Theorem}[section]
\newtheorem{lemma}[theorem]{Lemma}
\newtheorem{prop}[theorem]{Proposition}
\newtheorem{cor}{Corollary}
\newtheorem{definition}{Definition}[section]

\newcites{Supp}{Supplementary References}
\newcommand{\makesupplementtitle}[0]{\hsize\textwidth
    \linewidth\hsize \toptitlebar {\centering
        {\Large\bfseries Supplementary Material \par}}
    \bottomtitlebar}

\begin{document}

\runningtitle{\scriptsize Assessing Neural Network Representations During Training Using Noise-Resilient Diffusion Spectral Entropy}
\runningauthor{\scriptsize Assessing Neural Network Representations During Training Using Noise-Resilient Diffusion Spectral Entropy}

\twocolumn[

\aistatstitle{Assessing Neural Network Representations During Training Using Noise-Resilient Diffusion Spectral Entropy}

\aistatsauthor{
Danqi Liao$^{*}$\\Yale University \And
Chen Liu$^{*}$\\Yale University \And
Benjamin W. Christensen\\Yale University \AND
Alexander Tong\\Université de Montréal\\Mila -- Quebec AI Institute \And
Guillaume Huguet\\Université de Montréal\\Mila -- Quebec AI Institute \And
Guy Wolf\\Université de Montréal\\Mila -- Quebec AI Institute \AND
Maximilian Nickel\\Meta AI Research (FAIR) \And
Ian Adelstein\\Yale University \And
Smita Krishnaswamy~\Envelope\\Yale University\\Meta AI Research (FAIR)
}

\vskip 12pt
\aistatsaddress{
$^{*}$Equal contribution. Order of co-first authors determined by coin toss.\\
\Envelope~Please direct correspondence to \url{smita.krishnaswamy@yale.edu}
}



]

\begin{abstract}

Entropy and mutual information in neural networks provide rich information on the learning process, but they have proven difficult to compute reliably in high dimensions. Indeed, in noisy and high-dimensional data, traditional estimates in ambient dimensions approach a fixed entropy and are prohibitively hard to compute. To address these issues, we leverage data geometry to access the underlying manifold and reliably compute these information-theoretic measures. Specifically, we define diffusion spectral entropy (DSE) in neural representations of a dataset as well as diffusion spectral mutual information (DSMI) between different variables representing data. First, we show that they form noise-resistant measures of intrinsic dimensionality and relationship strength in high-dimensional simulated data that outperform classic Shannon entropy, nonparametric estimation, and mutual information neural estimation (MINE). We then study the evolution of representations in classification networks with supervised learning, self-supervision, or overfitting. We observe that (1) DSE of neural representations increases during training; (2) DSMI with the class label increases during generalizable learning but stays stagnant during overfitting; (3) DSMI with the input signal shows differing trends: on MNIST it increases, while on CIFAR-10 and STL-10 it decreases. Finally, we show that DSE can be used to guide better network initialization and that DSMI can be used to predict downstream classification accuracy across 962 models on ImageNet. The official implementation is available at~\url{https://github.com/ChenLiu-1996/DiffusionSpectralEntropy}.
\end{abstract}

\section{Introduction}

Deep neural networks have emerged as a major breakthrough in data science, mainly because of their ability to learn increasingly meaningful representations of data. Neural networks function by transforming data through a series of nonlinear operations, so that each layer learns a new representation of the data. Although the representation vectors reside in high-dimensional spaces, they are in fact located in lower-dimensional manifolds~\cite{ManifoldHypothesis}. Assessing the properties of this manifold therefore is key to better understanding the neural network.

The information bottleneck principle~\cite{InformationBottleneck} proposed a framework to measure mutual information between layers within a neural network. While this sheds light on the inner workings of neural networks, these quantities are historically difficult to compute~\cite{est_entropy_and_MI} and often fail when the data dimension increases~\cite{efficient_est_mi, MINE}, which we refer to as the curse of dimensionality. Specifically, as the dimension $D$ increases, the number of bins in a histogram used to calculate the probabilities increases exponentially in $D$, that is, $b^D$ where $b$ is the number of bins in each dimension. Indeed, recent investigations along this line are often restricted to small toy models: e.g., in~\cite{OnInformationBottleneck,OnInformationBottleneck2}, most experiments were performed on a 7-layer network of width 12-10-7-5-4-3-2.

Here, we use a powerful manifold learning paradigm, diffusion geometry, to present an alternative approach to quantifying entropy and mutual information. Diffusion geometry is known for its ability to access the underlying manifold of data representations~\cite{DiffusionMaps}, a property that is highly suitable for neural networks. With diffusion geometry, we can separate the noise dimensions from the signal dimensions and compute the entropy within the intrinsic dimensions of the manifold. By defining entropy on the eigenspectrum of the diffusion matrix, the problems incurred by the prohibitively high spatial dimension are mitigated. Further, by powering the matrix, we are able to low-pass filter the eigendimensions of the data, i.e., count noise dimensions with small eigenvalues proportionally less than dimensions with larger eigenvalues that explain more variance in diffusion coordinates.


A key contribution of this work is introducing diffusion spectral entropy~(DSE), or spectral entropy of the diffusion operator, as a robust quantifier of the intrinsic information measure of the representation of data in the presence of noise. Furthermore, we extend the diffusion spectral entropy to a diffusion spectral mutual information~(DSMI) in order to ascertain the information the embedding manifold has on the class labels or the raw input data from the dataset.

Our main contributions include the following.
\begin{itemize}
    \item Introducing \textit{diffusion spectral entropy (DSE)}, i.e., entropy of the diffusion operator spectrum as a measure of information in a data representation. Defining \textit{diffusion spectral mutual information (DSMI)} to assess the relationships between different layers of information in a neural network.
    \item Demonstrating that DSE and DSMI remain descriptive on very high-dimensional data and hence are suitable to modern-sized neural networks.
    \item Utilizing DSE and DSMI to assess the evolution of neural representations during training. 
    \item Demonstrating two ways to utilize DSE and DSMI. Specifically, using DSE to guide network initialization and using DSMI to predict downstream performance without full evaluation.
\end{itemize}

\section{Background}
\subsection{Manifold learning and diffusion geometry}

A useful assumption in representation learning is that high-dimensional data, commonly used in deep learning, originate from an intrinsic low-dimensional manifold that is mapped via nonlinear functions to observable high-dimensional measurements. This is commonly known as \textit{the manifold assumption}~\cite{ManifoldHypothesis}. Let $\mathcal{M}^d$ be a hidden $d$ dimensional manifold that is only observable via a collection of $n \gg d$ nonlinear functions $f_1,\ldots,f_n : \mathcal{M}^d \to \mathbbm{R}$ that enable its immersion in a high dimensional ambient space as $F(\mathcal{M}^d) = \{\mathbf{f}(z) = (f_1(z),\ldots,f_n(z))^T : z \in \mathcal{M}^d \} \subseteq \mathbbm{R}^n$ from which data are collected. Conversely, given a dataset $X = \{x_1, \ldots, x_N\} \subset \mathbbm{R}^n$ of high-dimensional observations, manifold learning methods assume that the data points originate from a sampling $Z = \{z_i\}_{i=1}^N \in \mathcal{M}^d$ of the underlying manifold through $x_i = \mathbf{f}(z_i)$, $i = 1, \ldots, n$ and aim to learn a low-dimensional intrinsic representation that approximates the manifold geometry of~$\mathcal{M}^d$.

A paradigm that has emerged in manifold learning in recent years is diffusion geometry~\cite{DiffusionMaps, magic, MELD, huguet2022time}. It models data based on transition or
random walk probabilities through the data and has been shown to be noise-tolerant. It has produced methods such as tSNE~\cite{tSNE}, PHATE~\cite{PHATE} and diffusion maps \cite{DiffusionMaps}. The geometry of the manifold can be learned with data diffusion by first computing the local similarities defined via a kernel $\mathcal{K}$. A popular choice is a Gaussian kernel $e^{-\|z_1 - z_2\|^2 / \sigma}$, where $\sigma > 0$ is interpreted as a user-configurable neighborhood size. To construct a diffusion geometry that is robust to sampling density variations, one would typically use an anisotropic kernel


\begin{equation}
\begin{aligned}
\label{eqn:diffusion_kernel}
\mathcal{K}(z_1, z_2) &= \frac{\mathcal{G}(z_1, z_2)}{\|\mathcal{G}(z_1, \cdot)\|_1^{\alpha} \|\mathcal{G}(z_2,\cdot)\|_1^{\alpha}} \textrm{, where} \\
\mathcal{G}(z_1, z_2) &= e^{-\frac{\|z_1 - z_2\|^2}{\sigma}}    
\end{aligned}
\end{equation}

\noindent as proposed in~\cite{DiffusionMaps}. $0 \leq \alpha \leq 1$ controls the separation of geometry from density. Next, the similarities encoded by $\mathcal{K}$ are normalized to define the transition probabilities $p(z_1, z_2) = \frac{\mathcal{K}(z_1, z_2)}{\|\mathcal{K}(z_1, \cdot)\|_1}$ that are organized in an $n \times n$ row stochastic matrix


\begin{equation}
\label{eqn:diffusion}
\mathbf{P}_{i,j} = p(z_i, z_j) = \frac{\mathcal{K}(z_1, z_2)}{\|\mathcal{K}(z_1, \cdot)\|_1}
\end{equation}

\noindent which describes a Markovian diffusion process on the intrinsic geometry of the data. 

\subsection{Entropy and mutual information}
\paragraph{Entropy}
Entropy quantifies the amount of uncertainty or ``surprise'' when given the value of a random variable. If the variable is distributed with a distribution that has a spread-out probability mass, such as a uniform distribution, then the entropy is high. On the other extreme, if there is no uncertainty in the quantity of the variable, i.e., it is deterministic, then the entropy is 0. The Shannon entropy is computed as follows.

\begin{equation}
\label{eqn:shannon_entropy}
    H(X) = \mathbb{E}[- \log p(X)] = - \sum_{x \in X} p(x) \log p(x)
\end{equation}

The von Neumann entropy~\cite{vonNeumannEntropy} from the quantum information domain extends the entropy measure to the quantum mechanics context and operates on density matrices. If a density matrix $\rho$ has a set of eigenvalues $\{ \eta_i \}$, the von Neumann entropy is defined as $H(\rho) = -tr(\rho \log \rho) = - \sum_{i} \eta_i \log \eta_i$.

Here, von Neumann entropy is considered to be an extension of Gibbs entropy, which is a measure of the spread of a distribution on the microstates of a classical system. Classical systems can only exist in pure states or standard basis states. However, quantum systems can exist in superposition states, and depending on the distribution of superposition states, the stable or ground states can be redefined as the eigenfunctions of a density operator that describes the probabilities of superpositions. 

This notion has subsequently been extended to graph spectra in several works. These methods generally compute the entropy of normalized eigenvalues of a graph adjacency matrix and have been used in biology and other fields to compare graphs ~\cite{topologyvariantionsse, de2016statisticalgraphspectral, takahashi2012discriminatinggraphspectra, merbis2023complex, informationonconectivitynetwork}.

\paragraph{Mutual information}
Mutual information is defined as a function of entropy. There are many equivalent formulations of mutual information that are equivalent. The most useful formulation here is as the difference between the (unconditional) entropy of a variable and the entropy of a variable conditioned on the other variable. The conditional entropy is computed as a weighted sum over discrete categories.

\begin{equation}
\begin{aligned}
\label{eqn:mutual_information}
I(X; Y) &= H(X) - H(X|Y) \\
&= H(X) - \sum_i p(Y=y_i) H(X|Y=y_i)
\end{aligned}
\end{equation}

\subsection{The information bottleneck theory}
The information bottleneck theory~\cite{InformationBottleneck} views a neural network with sequentially placed layers as a multistage information compressor. If we denote the neural representation of any layer as $Z$, the input signals as $X$ and the ground truth labels as $Y$, the theory concludes that $I(Z;X)$ will decrease while $I(Z;Y)$ will increase during training. Additionally, the theory claims that the optimal representation can be achieved by minimizing $I(Z;X) - \beta I(Z;Y)$.

\section{Related Work}

In efforts to quantify information in neural networks, in~\cite{InformationBottleneck}, the authors binned the vectors along each feature dimension to form a probability distribution and computed the Shannon entropy and mutual information. The main limitation of their method is the curse of dimensionality in the binning process that makes it impractical to analyze layers with more than a dozen neurons, which lag modern deep neural networks by orders of magnitude~(see Supplement~\ref{supp:limitations_classic_shannon}).

Researchers proposed using kernel density estimators, Kraskov estimators, as well as other nonparametric or parametric estimators~\cite{OnInformationBottleneck, KDE_MI, kraskov2004estimating, estimator1, estimator2, estimator3, estimator4, estimator5, efficient_est_mi} for mutual information estimation, yet these methods require assumptions on the distributions of hidden layer activation or do not generalize well on high-dimensional data~\cite{MINE}.

Among them, the nonparametric estimation toolbox~\cite{NPEET, kraskov2004estimating} locally estimates the log probability density at each sample point and averages them for a holistic estimate. Mutual information neural estimation (MINE)~\cite{MINE} uses a neural network, optimized by gradient descent, to approximate a lower bound of mutual information. 
In contrast, our proposed method does not assume a specific distribution on hidden-layer activations and remains descriptive in high dimensions.

\section{Methods}

The main difficulty in entropy and mutual information quantification in neural networks is the prohibitively high dimensionality. Binning and quantization fail quickly when we increase the dimension of the data (Supplement~\ref{supp:limitations_classic_shannon}). We leverage concepts from diffusion geometry to access the underlying manifold and compute these quantities reliably at high dimensions.

\subsection{Our definitions using diffusion geometry}

While spectral entropy has often been used to measure entropy on graphs, it has not been used often to compute the entropy of data. Here, we make a particular choice to compute a data-centric affinity matrix and then a spectral entropy from that matrix. In particular, we utilize the anisotropically normalized diffusion operator from Eqn~\ref{eqn:diffusion}.

We define the symmetric matrix $K$ by organizing the anisotropic kernel $\mathcal{K}$ in Eqn~\ref{eqn:diffusion_kernel} with $\alpha=\frac{1}{2}$
\begin{equation}
    \label{eqn:symmetric_kernel}
    K_{i, j} = \mathcal{K}(z_i, z_j)
\end{equation}

The row stochastic matrix $\mathbf{P}$ computed using Eqn~\ref{eqn:diffusion} with this $K$ is our diffusion matrix/operator.

When we compute the diffusion operator on the dataset $X$, we utilize the additional notation $\mathbf{P}_X$.

We define diffusion spectral entropy, with respect to a particular value of diffusion time $t$ as follows.

\begin{definition}
We define \textbf{diffusion spectral entropy (DSE)} as an entropy of the eigenvalues of the diffusion operator $\mathbf{P}_X$ calculated on a dataset $X$ where $x \in X$ is a multidimensional vector $[x_1, x_2 \ldots x_d]^T$: 

\begin{equation}
\label{eqn:diffspectralentropy}
S_D(\mathbf{P}_X,t):=-\sum_i \alpha_{i,t} \log(\alpha_{i,t})
\end{equation}
where $\alpha_{i,t}:= \frac{|\lambda_i^t|}{\sum_j |\lambda_j^t|}$,  and $\{ \lambda_i \}$ are the eigenvalues of the diffusion matrix $\mathbf{P}_X$.
\end{definition}

In the matrix $\mathbf{P}$ each data point is encoded according to its probability of transition to every other data point if a random walk on the data is made. Thus, if a data point is disconnected or far away from others, then a random walk starting at the data point is likely to remain at the data point. In this setting, the eigenvectors of the diffusion operator are paths through the data that are stable states of the transition operator. Thus, the entropy of the transition operator can be measured over the eigenbasis diffusion operator. Since rows of this matrix can also be viewed as representations of the data, this is also a measure of the intrinsic dimensionality of the dataset. Note that the parameter $t$ that parameterizes the entropy also gives us the capability to separate the noise from the true entropy of the signal. As the value of $t$ increases, the eigenspectrum shifts towards the low-frequency eigenvectors (which move slowly over the graph) because the eigenvalues $|\lambda_i|<1$ diminish at a rate inversely proportional to their value when they are raised to a power $t$. In fact, $\mathbf{P}^t$ has identical eigenvectors and powered eigenvalues as $\mathbf{P}$, which achieves a low-pass filtering of data values on the affinity graph~\cite{magic}. We note that a similar measure was used in the supplement of \cite{PHATE} to select parameters, but the diffusion spectral entropy was not defined or discussed there.

\begin{figure}[!hbt]
    \centering
    \includegraphics[width=0.48\textwidth,height=0.38\textwidth]{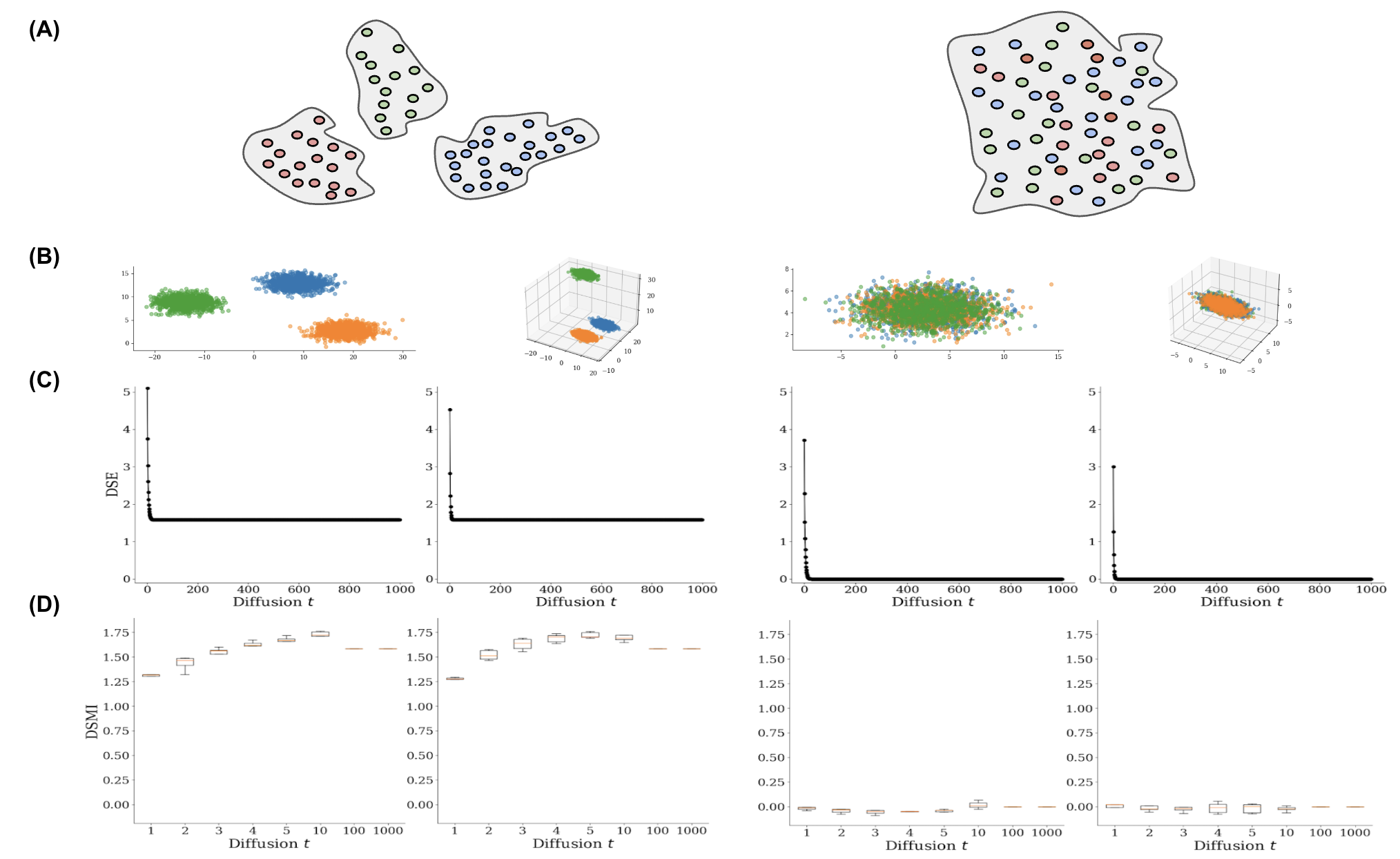}
    \caption{
       \textbf{Intuition for DSE and DSMI.} $X$: data, $Y$: class label. Colors represent classes. (A) sketch, (B) simulation, (C) DSE, (D) DSMI. In all panels, \textbf{Left:} $X$ forms $k$ clusters and $Y$ corresponds to clusters. $S_D(\mathbf{P}_X,t) \approx \log(k)$ at sufficiently large $t$ and $I_D(X; Y) > 0$. \textbf{Right:} $X$ forms a single cluster and $Y$ is randomly distributed over $X$. $S_D(\mathbf{P}_X,t) \approx 0$ at sufficiently large $t$ and $I_D(X; Y) \approx 0$.
       } \label{fig:intuition}
\end{figure}

An important remark is that, since DSE is computed on the eigenspectrum of the diffusion geometric representation of the data manifold, diffusion spectral entropy essentially counts information in terms of the number of eigendirections with non-trivial eigenvalues. In other words, \textit{DSE is more sensitive to the underlying dimension and structures (e.g., number of branches or clusters) than to the spread or noise in the data itself, which is contracted to the manifold by raising the diffusion operator to the power of $t$.} For an illustration of this phenomenon, see Figure~\ref{fig:intuition}(C), where as $t$ increases, the DSE converges to the log of number of distinct structures in the data. This is a highly valuable trait in the analysis of neural network representations, which often occurs on very noisy data. Indeed, neural representations that classify well create clear separations of the data into different eigendirections.

We further extend the diffusion spectral entropy to define mutual information for understanding the information that some variables of a data representation have on others, for example the information that neurons in a hidden layer have about the primary output.

\begin{definition}
\label{def:DSMI}
We define \textbf{Diffusion Spectral Mutual Information~(DSMI)} as the difference between unconditional and conditional diffusion spectral entropy, similar to the definition in Eqn~\ref{eqn:mutual_information}:

\begin{equation}
\begin{aligned}
\label{eqn:DSMI}
I_D(X; Y) =& S_D(\mathbf{P}_X,t) \\&- \sum_{y_i \in Y} p(Y=y_i) S_D(\mathbf{P}_{X|Y=y_i}, t)
\end{aligned}
\end{equation}
\end{definition}

$\mathbf{P}_{X|Y=y_i}$ is the transition matrix computed on the subset of $X$ that has class label $Y$. To avoid the numeric issues involved in comparing spectra of different sizes of matrices, we also computed $S_D(\mathbf{P}_X,t)$ after subsampling $X$ to the size of each class in $Y$. Since uniform subsampling maintains distributions, the sampled entropy would be the same as the total entropy as shown in our experiments~(see Supplement~\ref{supp:DSMI_additional_results}). 

\begin{figure*}[!htb]
    \centering
    \includegraphics[width=0.9\textwidth]{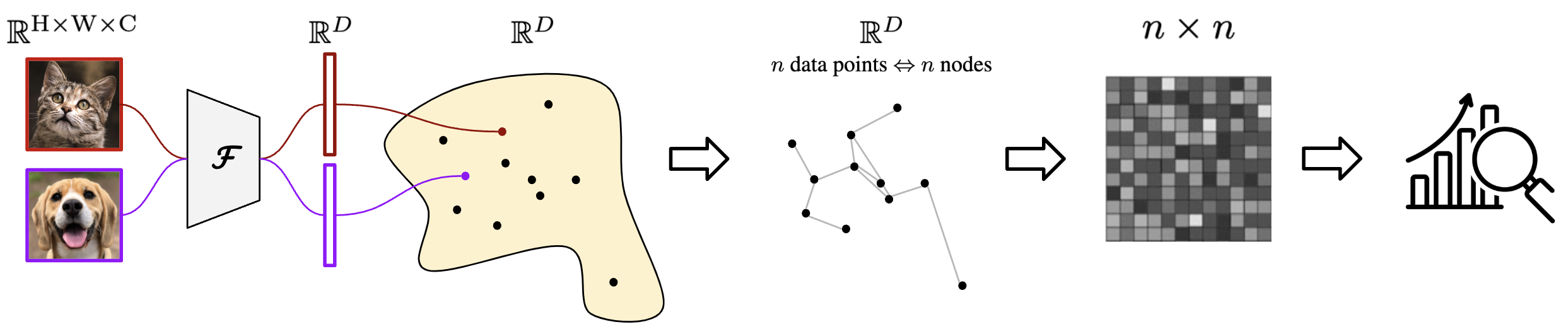}
    \caption{\textbf{Data processing to obtain the diffusion matrix of an embedding manifold.} Each data point is embedded by the network as a vector on a $D$-dimensional embedding manifold. This yields a point cloud that can be converted into a graph. We can compute its diffusion matrix, which allows for further analysis.}
  \label{fig:procedure}
\end{figure*}

Figure~\ref{fig:intuition} provides some insight into DSE and DSMI. In cases where the dataset $X$ forms $k$ clusters and the class label coincides exactly with well-separated clusters, DSE approaches $\log(k)$ with increasing $t$, because it counts the number of structures of the manifold instead of the data variance in the ambient space~(panel (C) left). For DSMI, based on Proposition \ref{prop:one-to-k}, the unconditional entropy with multiple clusters shall be greater than the conditional entropy within a single cluster, so DSMI will be positive~(panel (D) left). In the second case, where $X$ forms a single cluster and class labels are evenly distributed across the manifold, DSE approaches $\log(1)=0$ with increasing $t$ as there is only one structure~(panel (C) right). For DSMI, the conditional entropy $S_D(X|Y)$ and unconditional entropy $S_D(X)$ would be similar. As a result, DSMI shall be close to zero or even slightly negative for numerical reasons but, nonetheless, indicates low MI~(panel (D) right). See Supplement~\ref{supp:DSE_additional_results}~and~\ref{supp:DSMI_additional_results} for additional results.



\subsection{Properties and propositions}
Here we discuss some properties of DSE and DSMI, with \textbf{proofs in Supplement~\ref{supp:proposition_proofs}}. First, we provide the lower bound and the upper bound of $S_D$ when $t \rightarrow \infty$, and we explain the conditions when they are reached.

\begin{prop}
\label{prop:minimal_entropy}
$S_D$ achieves a minimal entropy of $0$ when the diffusion operator defines an ergodic Markov chain, and is in steady state (as $t \rightarrow \infty$). 
\end{prop}

Note that this also implies that if all data points are very similar and all points can easily transition to any other point, then it has minimal entropy. 



\begin{prop}
\label{prop:cluster_entropy}
As $t \rightarrow \infty$, $S_D(\mP_X, t)$ on data with $k$ well-separated clusters is $\log(k)$. 
\end{prop}

This shows that $S_D$ will reach its maximum value when the points are spread out very far apart.



Next, we examine the expected value of $S_D$.

\begin{prop}\label{prop:random-gaussian}
    Let $X \in \mathbb{R}^{n\times d}$ be a dataset of $n$ independent and identically distributed multivariate Gaussian vectors in $\mathbb{R}^d$, where $x_i \sim \mathcal{N}(0, I_d)$. Then, using $K$ as defined in Eqn~\ref{eqn:diffusion_kernel} with $\alpha = 1/2$, 

    \begin{equation*}
    \begin{aligned}
    &\mathbb{E}[S_D(\mathbf{P}_X, t = 1)] \\\lessapprox& \log(\frac{n}{1-\beta}) - \left(\frac{1}{n} + \left(\frac{n-1}{n}\right)\beta\right) \log\left(1+\frac{\beta n}{1 - \beta}\right)\\
    &\mathrm{where} \hspace{4pt} \beta = \left(1 + \frac{4}{\sigma}\right)^{-\frac{d}{2}}
    \end{aligned}
    \end{equation*}    
\end{prop}

This establishes a theoretical upper bound\footnote{It is also worth noting that the above bound remains valid for any distribution where $\|x_i - x_j\|^2_2 \sim f$ for $i \neq j$, where $f$ has the real-valued moment generating function $M(t)$. Then, we simply set $\beta = M(-\frac{1}{\sigma})$.}
 on the DSE at any given layer. Since independence is not necessarily assumed in general, the DSE for any dataset in practice should be less than this bound. Furthermore, this bound intuitively reinforces that for large $d$, $\beta \approx 0$, so $\mathbb{E}[S_D(\mathbf{P}_X, t = 1)] \approx \log(n)$. 

Our last proposition on DSE relates DSE of a single cluster to that of multiple clusters. This has implications for classifier training.

\begin{figure*}[!hb]
  \begin{minipage}[c]{0.7\textwidth}
    \includegraphics[width=\textwidth]{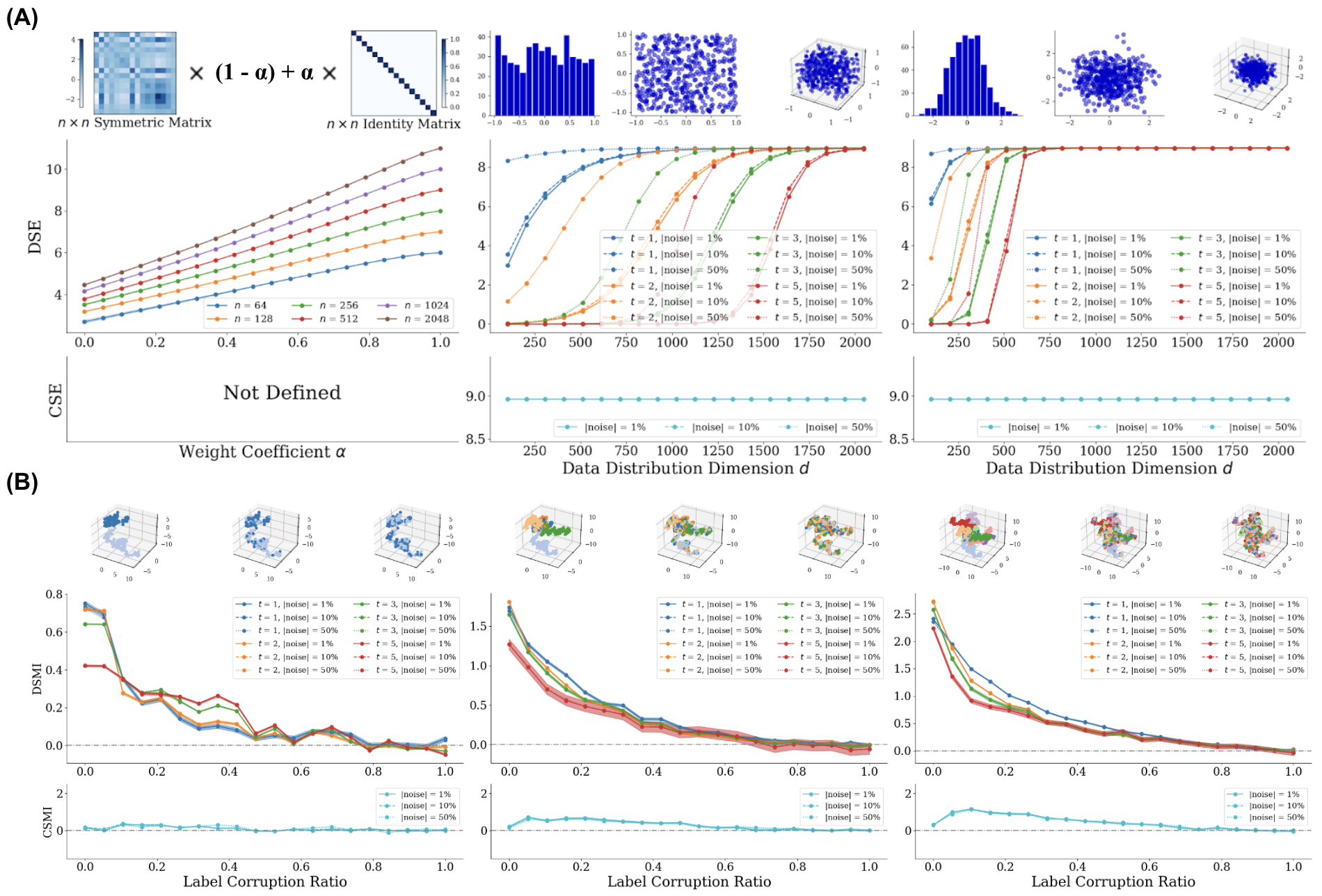}
  \end{minipage}\hfill
  \begin{minipage}[c]{0.28\textwidth}
    \caption{\textbf{Diffusion Spectral Entropy~(DSE) and Diffusion Spectral Mutual Information~(DSMI) on toy data.} \textbf{(A)} DSE increases as intrinsic dimension grows, while classic Shannon entropy~(CSE) saturates to $\log(n) = \log_2(500) = 8.966$ due to curse of dimensionality. \textbf{(B)} When two random variables are dependent, DSMI negatively correlates with the level of data corruption, while classic Shannon mutual information~(CSMI) does not capture this trend. See full description in Supplement~\ref{sec:full_caption_toy_data_entropy_MI}.}
  \label{fig:toy_data_entropy_MI}
  \end{minipage}
\end{figure*}

\begin{prop}\label{prop:one-to-k}
Take $n$ to be arbitrarily large. Let $X \in \mathbb{R}^{n\times d}$ be a matrix of i.i.d. random values $x_{ij} \sim f$. Let $Y \in \mathbb{R}^{n\times d}$ be a matrix of i.i.d. random values $y_{ij} \sim f$, but in $k \in \mathbb{N}$ distinct clusters such that when the anisotropic probability matrix is computed for $\alpha = 1/2$, the probability of diffusion between points of different clusters is arbitrarily small. Then, using $\beta$ as defined in Proposition \ref{prop:random-gaussian}, the approximate upper bound on DSE increases by $\beta\log(k)$.
\end{prop}

Recall the learning process of a classification network. As the model learns, it separates latent representations into distinct clusters throughout the process. Proposition~\ref{prop:one-to-k} elucidates that the expected upper bound of $S_D$ will increase during the training of a classifier.

DSMI has a minimal value of 0, achieved when the random variables are mutually independent. On the other hand, its maximal value can be bounded as follows.
\begin{prop}
\label{prop:cluster_mi}
As $t \rightarrow \infty$ in a hidden layer $X$ with $k$ well-separated clusters and class labels $Y$ perfectly indicating clusters, we will have $I_D(X;Y)=\log(k)$. 
\end{prop}

\subsection{Efficiently computing DSE and DSMI}

To compute DSE and DSMI on neural network representations, we pass the training data $X$ to the desired layer $L$ of a neural network. At this layer, we collect the activations of each neuron into a vector $L(x_i) = [L_1(x_i), \ldots, L_n(x_i)]$, where $L_j$ is the $j$-th neuron of the $L$-th layer. $L(x_i)$ is a $n$-dimensional representation. If the activation is a multidimensional tensor, we can flatten it into a vector. We then compute $\mathbf{P}_{L(X)}$ and proceed to compute DSE and DSMI. This process is illustrated in Figure~\ref{fig:procedure}.

The calculation of eigenvalues via complete eigendecomposition is known to have a time complexity $\mathcal{O}(n^3)$. However, we take advantage of two characteristics to provide a faster method: 1) we do not need the eigenvectors, 2) $\mathbf{P}$ has the same eigenvalues as the real symmetric matrix $K$ defined in Eqn~\ref{eqn:symmetric_kernel} as shown in~\cite{DiffusionMaps}. Thus, we can compute the DSE efficiently using QR decomposition since we only need the eigenvalues of a real symmetric matrix. 

\begin{figure*}[!bt]
  \begin{minipage}[c]{0.7\textwidth}
    \includegraphics[width=\textwidth]{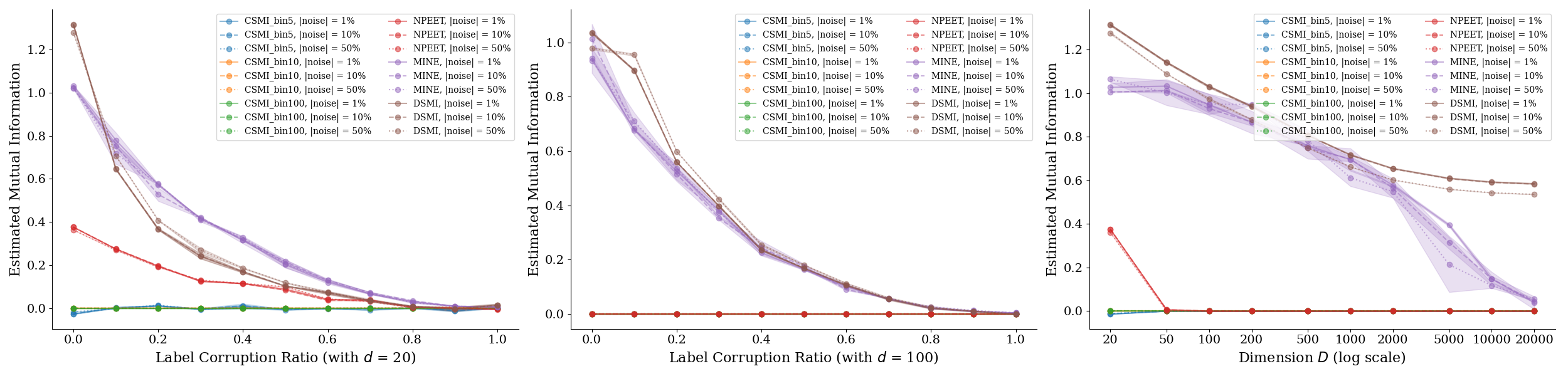}
    \caption{Mutual information estimation on Gaussian blobs.}
    \label{fig:toy_data_MI_blob}
  \end{minipage}\hfill
  \begin{minipage}[c]{0.28\textwidth}
    \includegraphics[width=\textwidth]{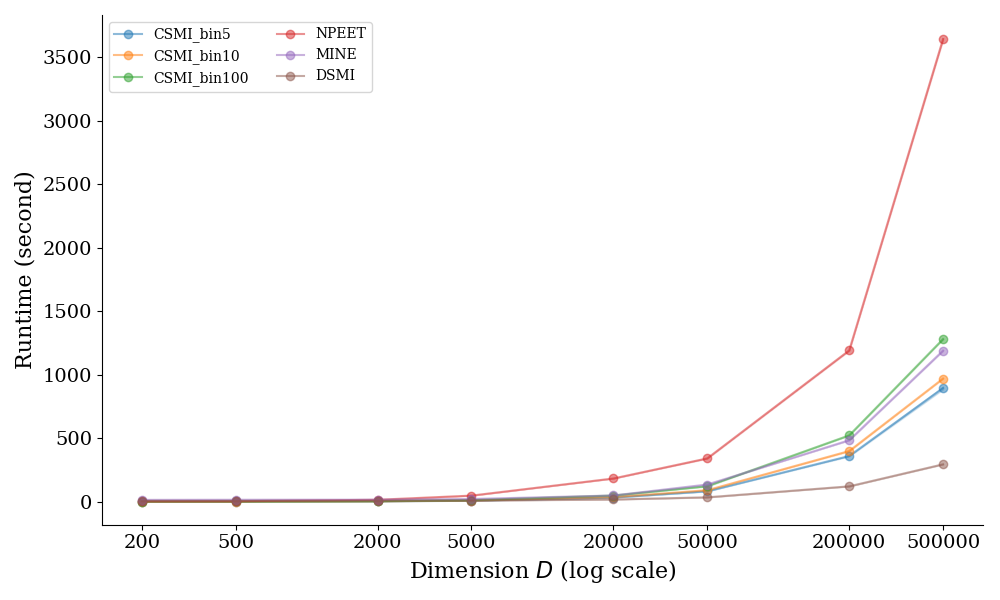}
    \caption{Runtime comparison.}
    \label{fig:runtime}
  \end{minipage}
\end{figure*}

\section{Results}

\subsection{Toy test cases for DSE and DSMI}

To demonstrate the behavior of DSE, we first performed several simulations as shown in Figure~\ref{fig:toy_data_entropy_MI}(A). The left panel indicates that, for an arbitrary real symmetric matrix, the closer it gets to an identity matrix, the higher its DSE --- since an $n \times n$ identity matrix represents $n$ distinct clusters/states. The two other panels show that DSE in general increases as the intrinsic dimensionality of the data manifold increases. 

To demonstrate the behavior of DSMI, we performed similar simulations as shown in Figure~\ref{fig:toy_data_entropy_MI}(B). In these toy test cases, we gradually corrupt the class label on the simulated tree branches. Just as expected, DSMI starts high when the association between the data point and the class label is high~(i.e., when each branch is assigned a distinct label), and drops to zero as the corruption gradually brings the label to complete noise. 

We also computed the classic Shannon entropy~(CSE) and classic Shannon mutual information~(CSMI) for comparison using the method from~\cite{InformationBottleneck}. Our proposed method outperforms the classic Shannon version in high-dimensional spaces. In Figure~\ref{fig:toy_data_entropy_MI}(A), DSE consistently captures entropy trends while CSE saturates to $\log(n)$. In Figure~\ref{fig:toy_data_entropy_MI}(B), DSMI follows the expected trend, while CSMI does not.

\subsection{DSMI at very high dimensions}
\label{sec:dsmi_scale_dim}

In this section, we show how DSMI better scales to high dimensions compared to existing methods. We ran mutual information estimators (DSMI, CSMI, a nonparametric entropy and mutual information estimation toolbox (NPEET)~\cite{NPEET, kraskov2004estimating} and mutual information neural estimation (MINE)~\cite{MINE}) on labeled Gaussian blobs of various dimensions. The left and middle panels of Figure~\ref{fig:toy_data_MI_blob} show that all methods generally obey the expected behavior as they show positive mutual information when the label is clean and gradually degrade to zero when the label is fully corrupted. The right panel indicates that CSMI, NPEET, and MINE fail as the dimension gets large, while DSMI remains significant.

Moreover, DSMI scales better in runtime~(Figure~\ref{fig:runtime}). This is expected since the time complexity for DSMI is $\mathcal{O}(n^2 \cdot D + n^3)$ due to the cost of pairwise distance computation and eigendecomposition (without considering optimization), while that for CSMI is $\mathcal{O}(2^D)$ due to the cost of computing exponentially many data bins.

\subsection{Evolution of DSE and DSMI during training}
\label{sec:real_exp}

To analyze DSE and DSMI on real data, we trained common vision networks and assessed the penultimate layer of each network at the end of each epoch. To cover the variety of model architectures, we experimented with 3 convolutional neural networks and 3 vision transformers. See supplementary~\ref{supp:experiment_details} for details.

We specifically investigated the penultimate layer because it is usually believed that the representations in this layer reflect the learning of the entire network. As a reference, it is common practice to directly use the penaltimate layer to evaluate the overall representation power of the network, e.g., linear probing~\cite{SimCLR}. Although our experiments mainly focus on this layer, our evaluation framework can be easily adapted to other layers in the network.

We trained the vision backbones under three conditions: supervised learning, contrastive learning, and intentional overfitting on randomized nonsense labels, repeated under 3 random seeds. More details can be found in Supplement~\ref{supp:experiment_details}. For all of the DSE and DSMI experiments below, results of the classic Shannon version can be found in Supplement~\ref{supp:limitations_classic_shannon}, and the raw results for each network are included in Supplement~\ref{supp:all_raw_results}.

\paragraph{DSE of a hidden layer}

\begin{figure}[!tb]
    \centering
    \includegraphics[width=0.48\textwidth]{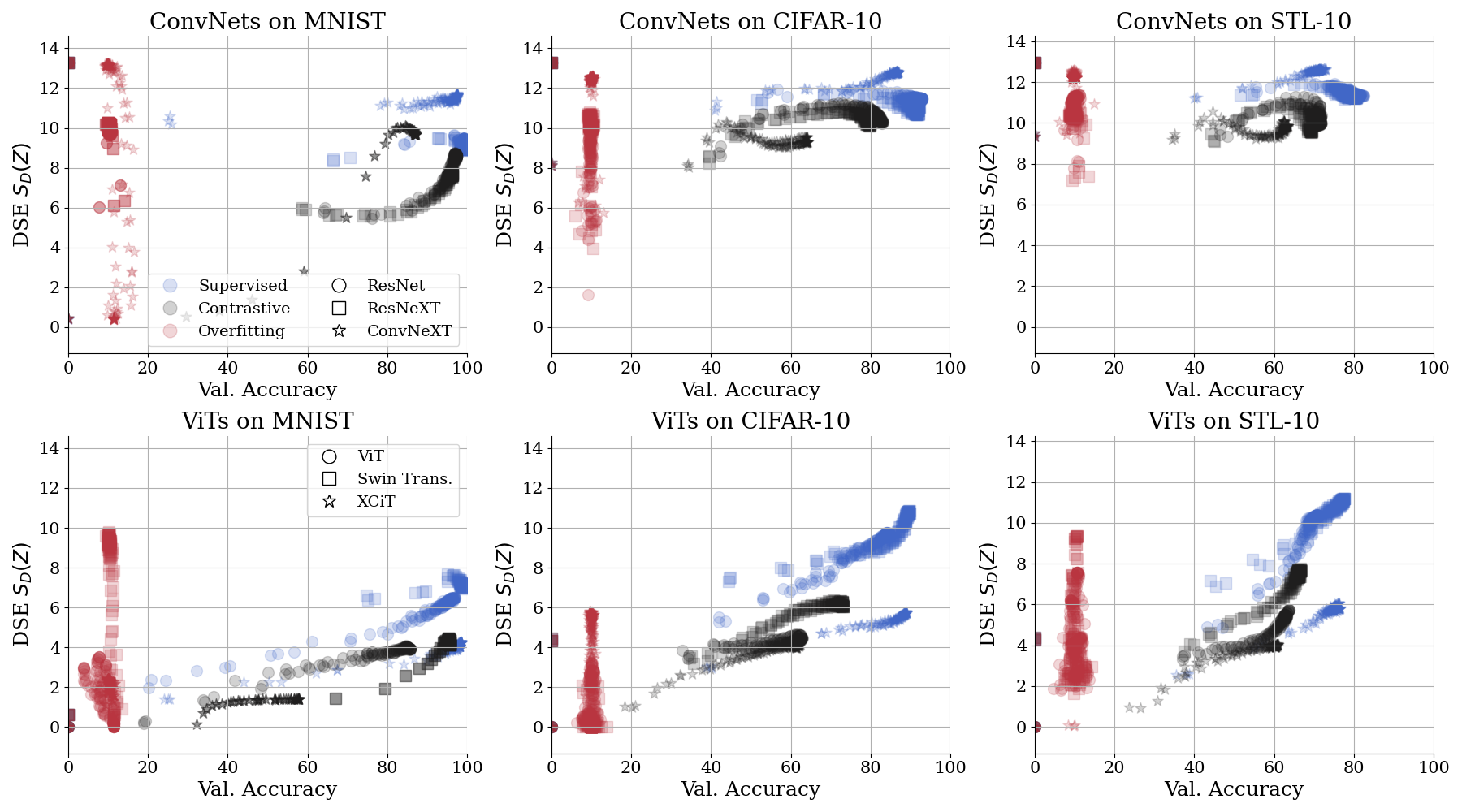}
    \caption{DSE $S_D(Z)$ of representation $Z$.}
    \label{fig:main_results_DSE}
\end{figure}

We can observe from Figure~\ref{fig:main_results_DSE} that DSE in proper learning (i.e. supervised or contrastive learning on correct labels) increases as the model performs better on the classification task. When the model is forced to memorize random nonsense labels, the entropy increases similarly even though the classification performance is stagnant.

\paragraph{DSMI of a hidden layer with the class label}

\begin{figure}[!tb]
    \centering
    \includegraphics[width=0.48\textwidth]{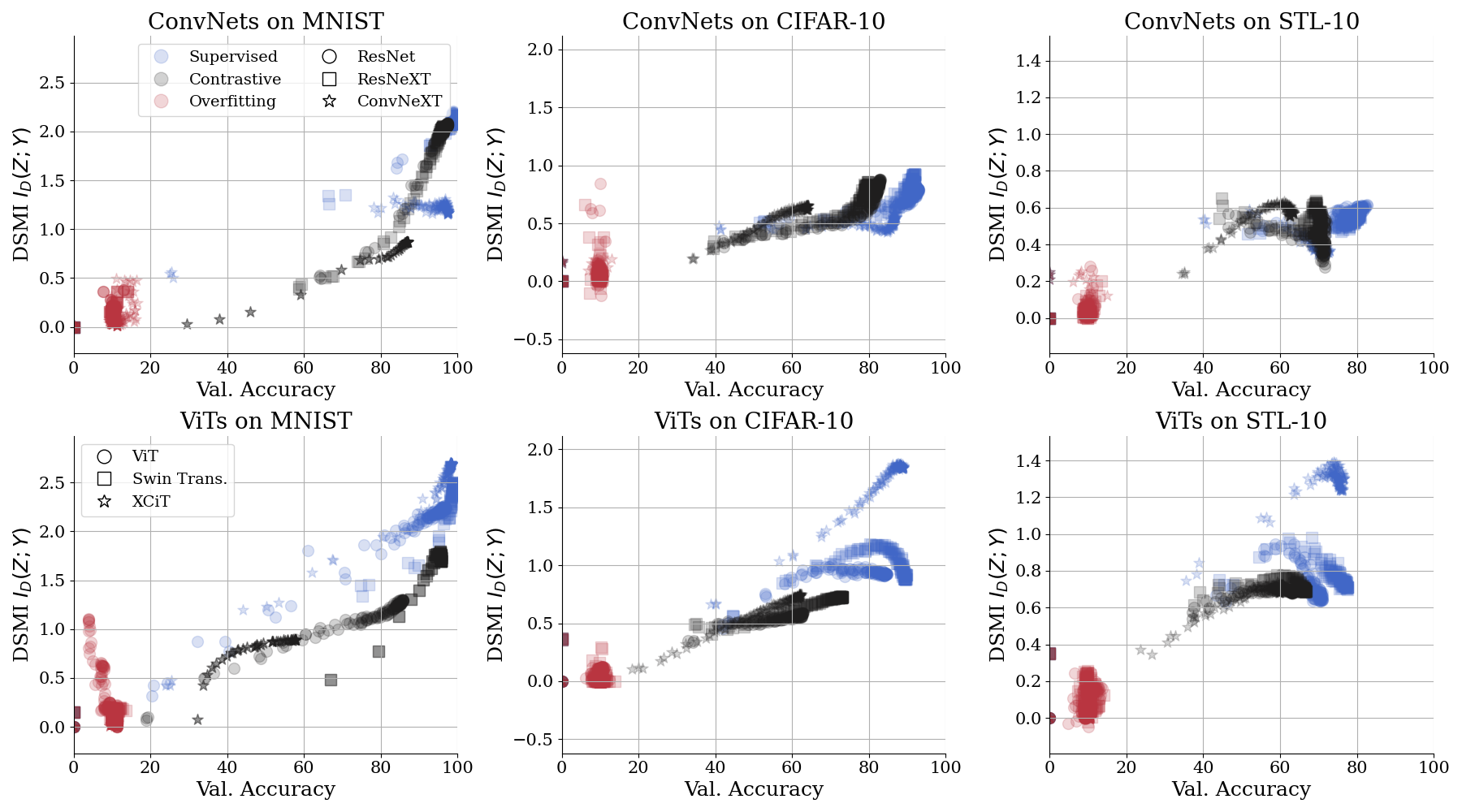}
    \caption{DSMI $I_D(Z; Y)$ between representation $Z$ and the class label $Y$.}
    \label{fig:main_results_DSMI_output}
\end{figure}

In Figure~\ref{fig:main_results_DSMI_output}, it can be seen that DSMI $I_D(Z; Y)$ increases consistently during proper learning. In many cases, DSMI climbs more slowly in contrastive learning compared to supervised learning and ends up at a lower terminal value. This may be attributed to the fact that contrastive learning lacks direct supervision from explicit class labels. However, since class labels relate to the data geometry, self-supervised learning on the data alone still yields some mutual information with labels. In nonsense memorization, DSMI quickly converges to around zero. This aligns well with the expectation, since a classifier that essentially performs random guessing has zero mutual information with the class label, whereas a functioning classifier corresponds to positive mutual information.

\begin{figure}[!tb]
    \centering
    \includegraphics[width=0.48\textwidth]{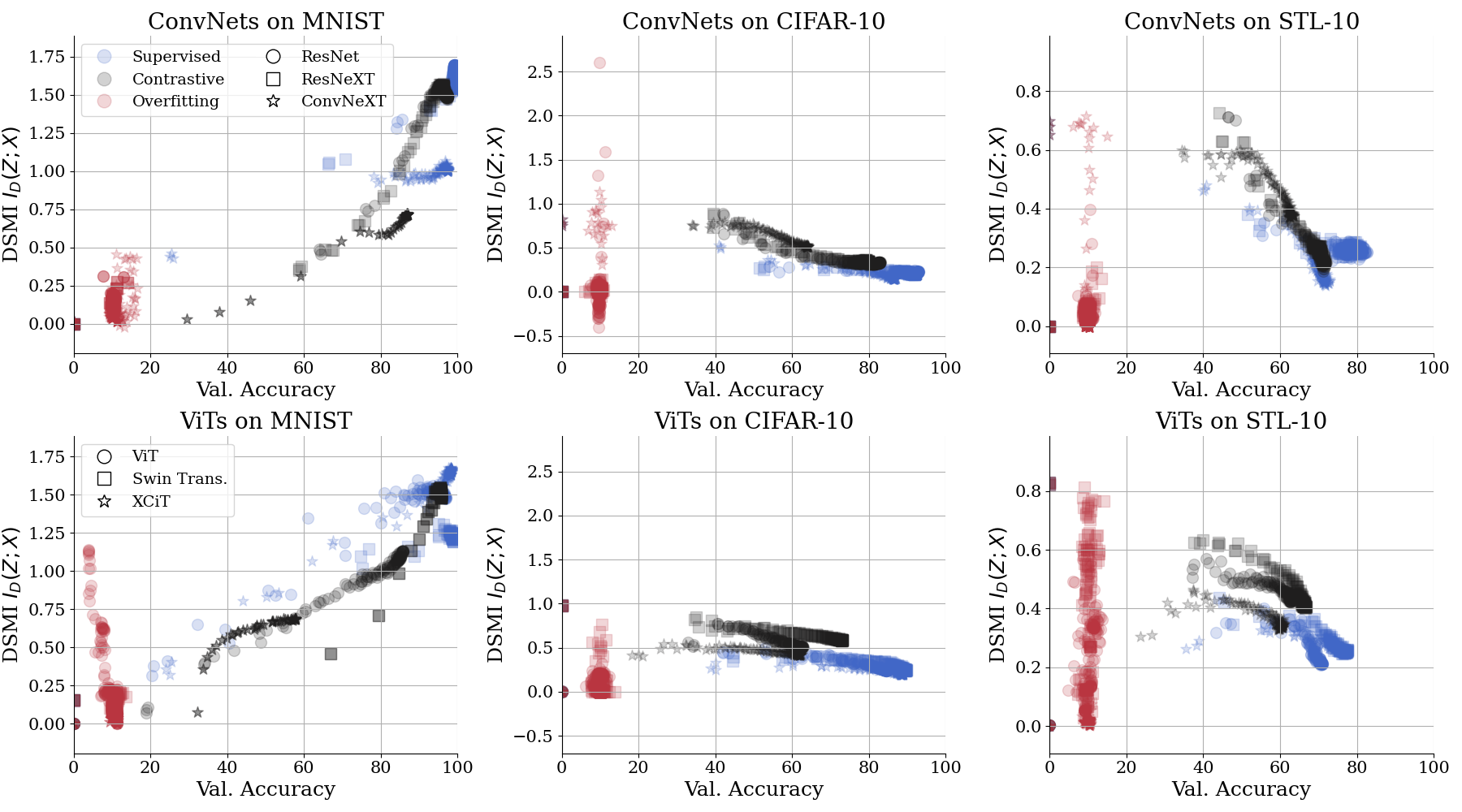}
    \caption{DSMI $I_D(Z; X)$ between $Z$ and input $X$.}
    \label{fig:main_results_DSMI_input}
\end{figure}


\paragraph{DSMI of a hidden layer with the input}

We also show DSMI with the input signal in Figure~\ref{fig:main_results_DSMI_input}. During proper learning, the information bottleneck theory would suggest $I_D(Z; X)$ should decrease~\cite{InformationBottleneck} while counterarguments have also been provided in~\cite{OnInformationBottleneck}. Our results suggest that both may be correct and the trend may depend on the nature of the dataset $X$. $I_D(Z; X)$ keeps increasing during learning on the MNIST dataset, while it mostly decreases after an initial increase on the CIFAR-10 and STL-10 datasets. We suspect that all models are essentially overfitting the MNIST data, and this may be verified by future studies. In nonsense memorization, $I_D(Z; X)$ rises to a significant level in most cases. This indicates that even when they overfit random labels, the neural representations still contain information about the input signals.

\subsection{Guiding network initialization}

We sought to assess the effects of network initialization in terms of DSE. We were motivated by two observations~(see Figure~\ref{fig:raw_results_DSE} in Supplement~\ref{supp:all_raw_results}): (1) the initial DSEs for different models are not always the same despite using the same method for random initialization; (2) if DSE starts low, it grows monotonically; if DSE starts high, it first decreases and then increases.

\begin{figure}[!thb]
    \begin{minipage}[c]{0.12\textwidth}
    \includegraphics[width=\textwidth]{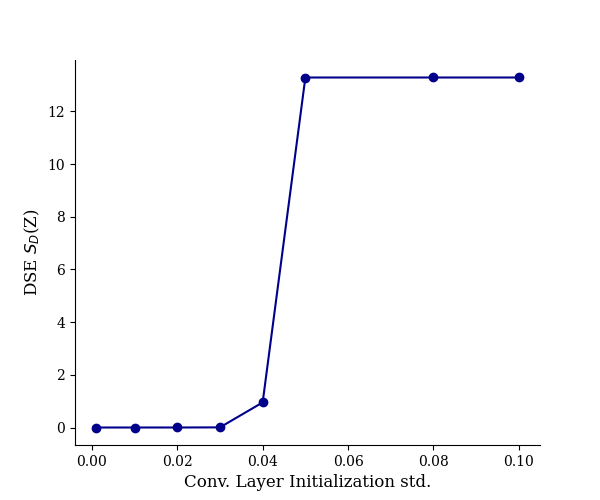}
    \end{minipage}\hfill
    \begin{minipage}[c]{0.37\textwidth}
    \includegraphics[width=\textwidth, height=0.33\textwidth]{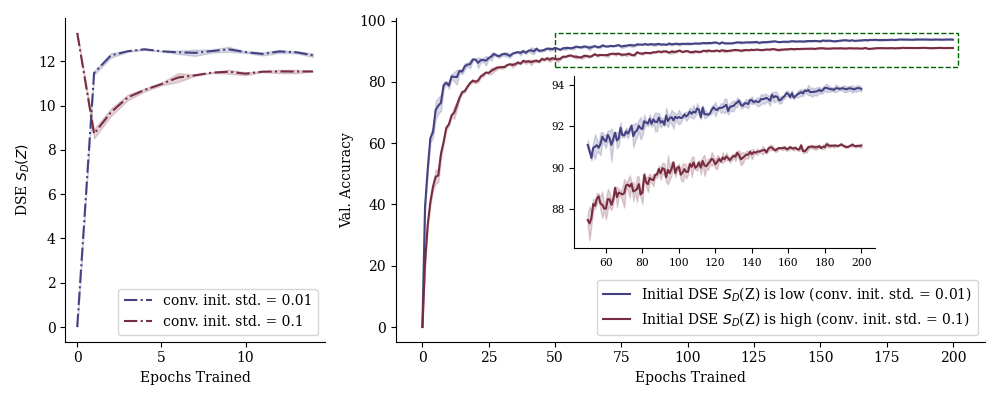}
    \end{minipage}

    \begin{minipage}[c]{0.12\textwidth}
    \includegraphics[width=\textwidth]{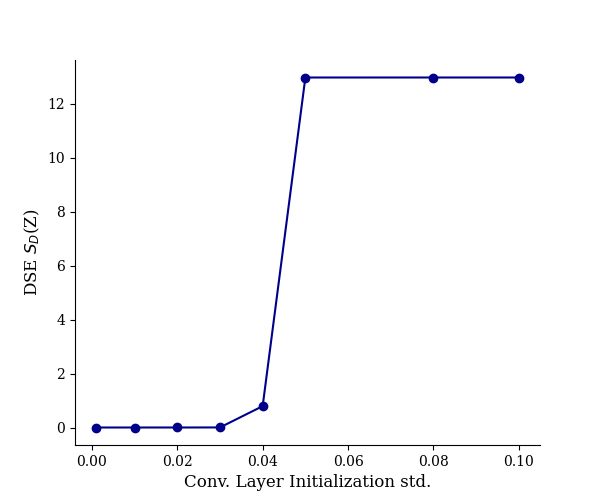}
    \end{minipage}\hfill
    \begin{minipage}[c]{0.37\textwidth}
    \includegraphics[width=\textwidth, height=0.33\textwidth]{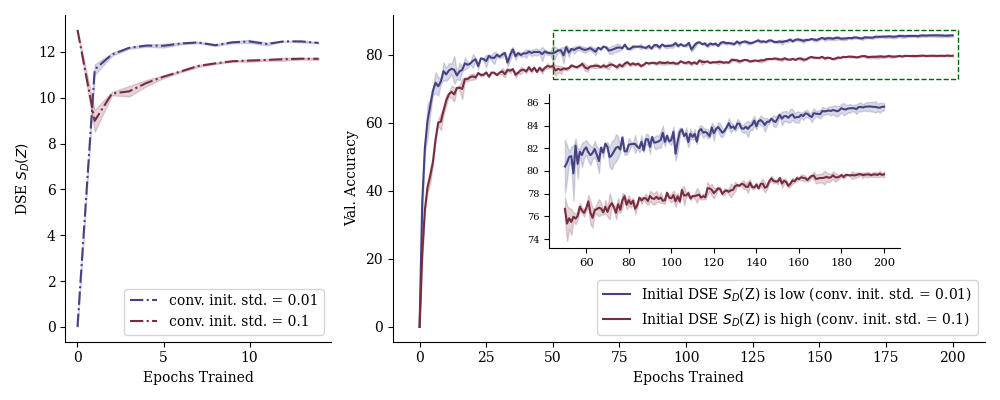}
    \end{minipage}

    \caption{Initializing the network at a low DSE allows for faster convergence and better final performance. Top: CIFAR-10, bottom: STL-10. Shaded areas indicate standard deviation across 3 random seeds.}
    \label{fig:DSE_init_compare}
\end{figure}

\begin{figure*}[!thb]
    \centering
    \includegraphics[width=0.94\textwidth]{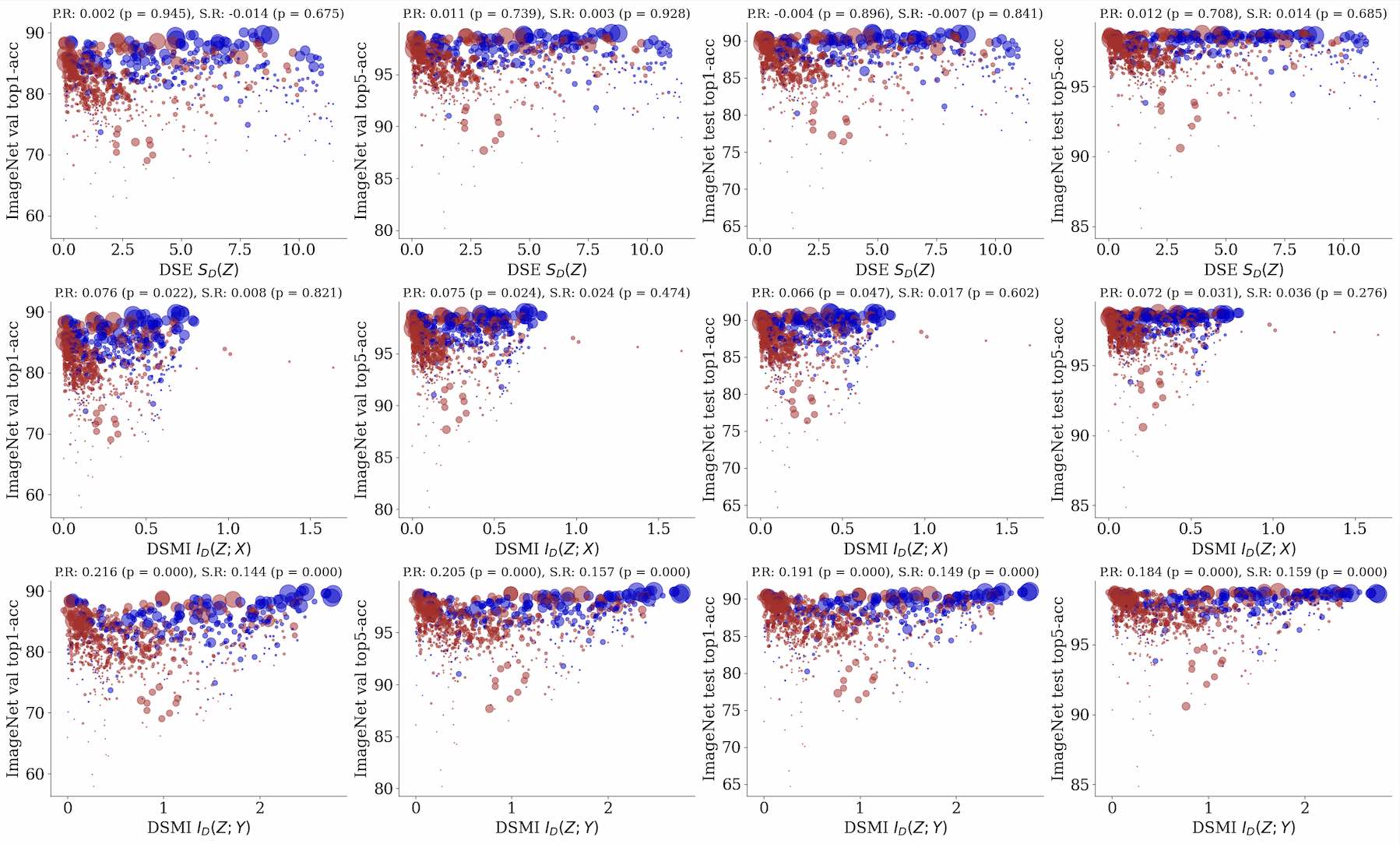}
    \caption{\textbf{Correlation analysis between DSE $S_D(Z)$, DSMI $I_D(Z; X)$, $I_D(Z; Y)$ and ImageNet accuracy evaluated on 962 pre-trained models.} Red circles are ConvNets and blue circles are ViTs. Circle sizes indicate model sizes. $I_D(Z; Y)$ shows a strong positive correlation ($p < 0.001$). P.R: Pearson correlation coefficient, S.R: Spearman correlation coefficient.}
    \label{fig:vs_imagenet_acc}
\end{figure*}

We found that if we initialize the convolutional layers with weights $\sim \mathcal{N}(0, \sigma)$, DSE $S_D(Z)$ is affected by $\sigma$~(Figure~\ref{fig:DSE_init_compare} left). We then trained ResNet models with networks initialized at high ($\approx \log(n)$) versus low ($\approx 0$) DSE by setting $\sigma=0.1$ and $\sigma=0.01$, respectively. The training history suggests that initializing the network at a lower $S_D(Z)$ can improve the convergence speed and final performance~(Figure~\ref{fig:DSE_init_compare}). We believe this is because the high initial DSE from random initialization corresponds to an undesirable high-entropy state, which the network needs to get away from (causing the DSE decrease) before it migrates to the desirable high-entropy state (causing the DSE increase). More details can be found in Supplement~\ref{supp:dse_initialization}.

\subsection{ImageNet cross-model correlation}

In addition to evaluating DSE and DSMI \textit{along the training process of the same model}, we investigated whether they can serve as meaningful indicators \textit{across different models}. To that end, we computed the correlations between these diffusion geometric quantities on a representative subset of data and the downstream classification accuracy across 962 publicly available models pre-trained on ImageNet. Details of this experiment are described in Supplement~\ref{supp:experiment_details}. As illustrated in Figure~\ref{fig:vs_imagenet_acc}, $I_D(Z; Y)$ has a strong positive correlation with classification accuracy (with $p < 0.001$ consistently under four conditions). On the other hand, $S_D(Z; X)$ and $I_D(Z; X)$ do not show strong statistical significance. This indicates a trend that a higher $I_D(Z; Y)$ corresponds to a better downstream performance, which may facilitate the selection of promising models without evaluating on the entire dataset.

\section{Conclusion}
In this work, we define diffusion spectral entropy (DSE) and diffusion spectral mutual information (DSMI). Through simulation on toy datasets, we showed that DSE and DSMI are meaningful and noise-resilient measures that can scale better to very high dimensions compared to existing methods in both reliability and runtime. We then analyzed the neural representations of six vision backbones in supervised learning, contrastive learning, and overfitting settings. We observed that DSE during learning increases under all training conditions. Furthermore, we showed that DSMI between a hidden layer and class labels increases during learning, whereas it converges to zero when overfitting random labels. We saw more complex trends in DSMI with the primary input: it increases in the MNIST dataset, while it decreases in CIFAR-10 and STL-10. Finally, we show that DSE can be used to guide network initialization and that DSMI is strongly correlated with downstream performance.


Some future directions include further investigating the effect of network initialization, exploring DSE and DSMI as regularizations for supervised learning, using DSE to regularize self-supervised learning, using DSMI to improve adversarially trained generative models, and many other exciting opportunities. Researchers can further extend this framework to data from other systems, in addition to neural networks, to understand how neural networks process information similarly or differently from other systems, such as brain networks.

\clearpage
\newpage
\bibliography{references}
\bibliographystyle{apalike}

\renewcommand{\thefigure}{S\arabic{figure}}
\renewcommand{\theHfigure}{S\arabic{figure}}
\renewcommand{\thesection}{S\arabic{section}}
\renewcommand{\theHsection}{S\arabic{section}}
\renewcommand{\thetable}{S\arabic{table}}
\renewcommand{\theHtable}{S\arabic{table}}
\setcounter{figure}{0}

\clearpage

\thispagestyle{empty}

\onecolumn \makesupplementtitle

\appendix


\renewcommand{\baselinestretch}{0.0}\normalsize
\startcontents[sections]
\printcontents[sections]{l}{1}{\setcounter{tocdepth}{2}}
\vskip 12pt
\renewcommand{\baselinestretch}{1.0}\normalsize

\vskip 0.1in
\section{Limitations of the Classic Shannon Entropy and Mutual Information}
\label{supp:limitations_classic_shannon}

\subsection{Computing the Classic Shannon Entropy and Mutual Information}

To compute the classic Shannon entropy of a variable $Z$, instead of computing the entropy on the eigenvalue spectrum of the diffusion matrix which gives $S_D(Z)$, we directly compute the entropy on the embedding vectors which give $H(Z)$.

For $n$ data points, the set of embedding vectors contains a total of $n$ vectors in $\mathbb{R}^D$. We need to convert the $n$ vectors to probability densities and use Eqn~\ref{eqn:shannon_entropy}. The most common way (e.g., in~\citeSupp{InformationBottleneck}) is to bin these vectors along each of the $D$ feature dimensions. Specifically, we will compute the global range of all $n$ vectors along each feature dimension $i \in \{1, 2, \ldots, D \}$, and normalize them to $[0, 1]$. Then, we can quantize all the vectors along each dimension into $b$ different bins. For example, if $b = 10$, values in $[0, 0.1)$ will be assigned to bin $1$; values in $[0.1, 0.2)$ to bin $2$, etc. As a result, each vector is converted to a quantized version, with each entry being an integer in $[1, b]$. Each possible quantized vector can be referred to as a ``bucket''. It can be easily noticed that the number of buckets is equal to $b^D$. After counting the number of vectors in each bucket, we can estimate the distribution of probability density on the buckets. Finally, we compute the classic Shannon entropy using Eqn~\ref{eqn:shannon_entropy}.  Classic Shannon mutual information can be computed in a similar manner, using Eqn~\ref{eqn:mutual_information} and the aforementioned entropy computation.

\subsection{Limitations of Classic Shannon Entropy and Mutual Information}

\begin{figure*}[!htb]
    \centering
    \includegraphics[width=0.75\textwidth]{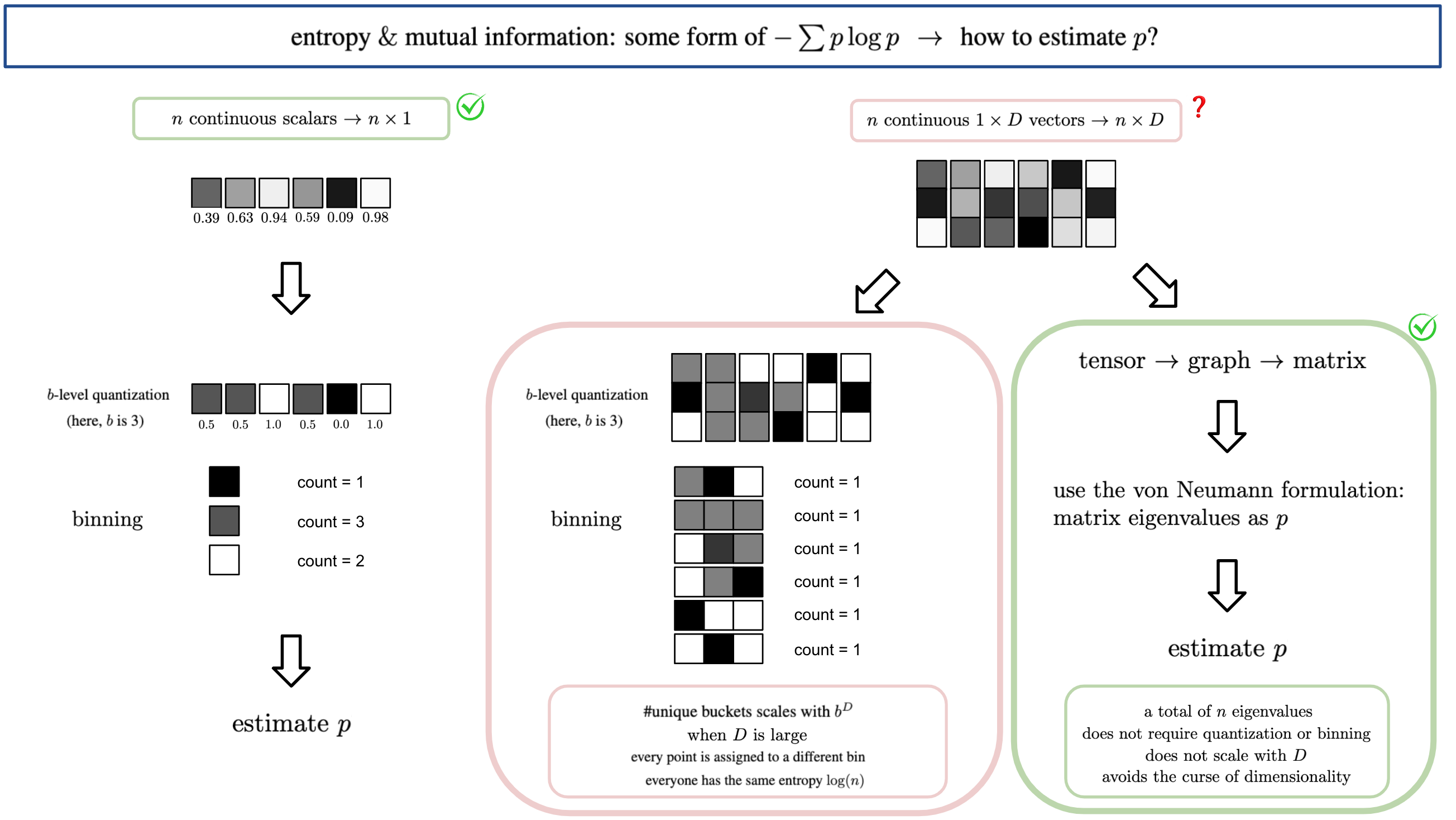}
    \caption{\textbf{Curse of dimensionality in Classic Shannon Entropy / Mutual Information.}}
    \label{fig:curse_of_dim}
\end{figure*}

The major limitation lies in the binning process. It is a known problem that \textbf{the number of buckets scales exponentially with respect to the feature dimension} $D$. Take ResNet-50 as an example: the penultimate layer has $D = 2048$. It is overwhelmingly likely that all embedding vectors are assigned to different buckets, even if we use the minimal choice of 2 bins per feature dimension --- which is already a very coarse-grained binning. When the majority of cases result in unique bucketing which leads to the maximum entropy, this metric has very limited expressiveness. This phenomenon, which we call the curse of dimensionality in the CSE and CSMI computation, is illustrated in Figure~\ref{fig:curse_of_dim}.

\begin{figure*}[!htb]
    \centering
    \includegraphics[width=0.7\textwidth]{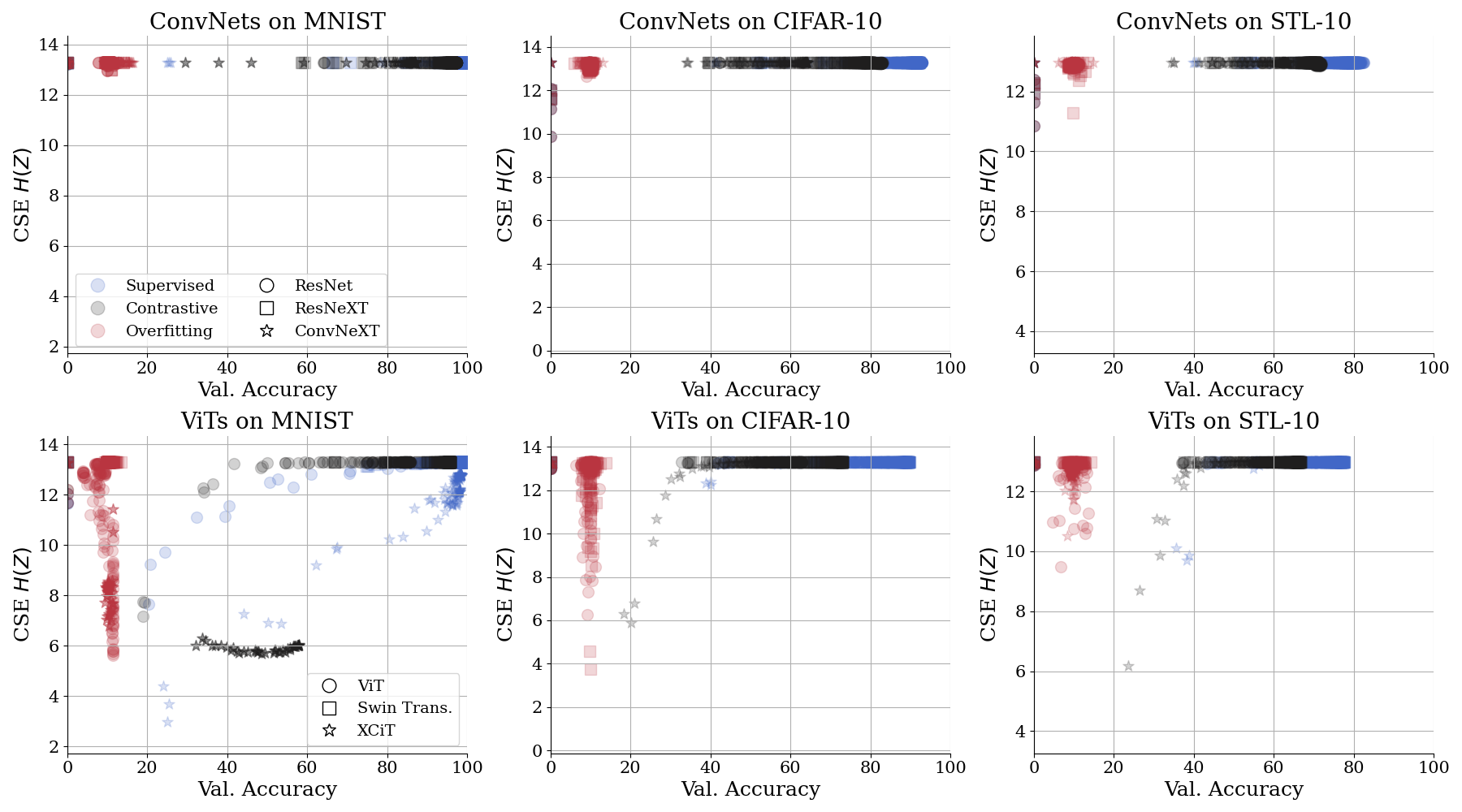}
    \caption{\textbf{Classic Shannon entropy version for Figure~\ref{fig:main_results_DSE}.} The embedding vectors are frequently allocated to unique buckets, which leads to the maximum possible entropy of $-\log_2 \frac{1}{10000} = 13.288$ for 10,000 data points. }
    \label{fig:main_results_CSE}
\end{figure*}

\begin{figure*}[!htb]
    \centering
    \includegraphics[width=0.7\textwidth]{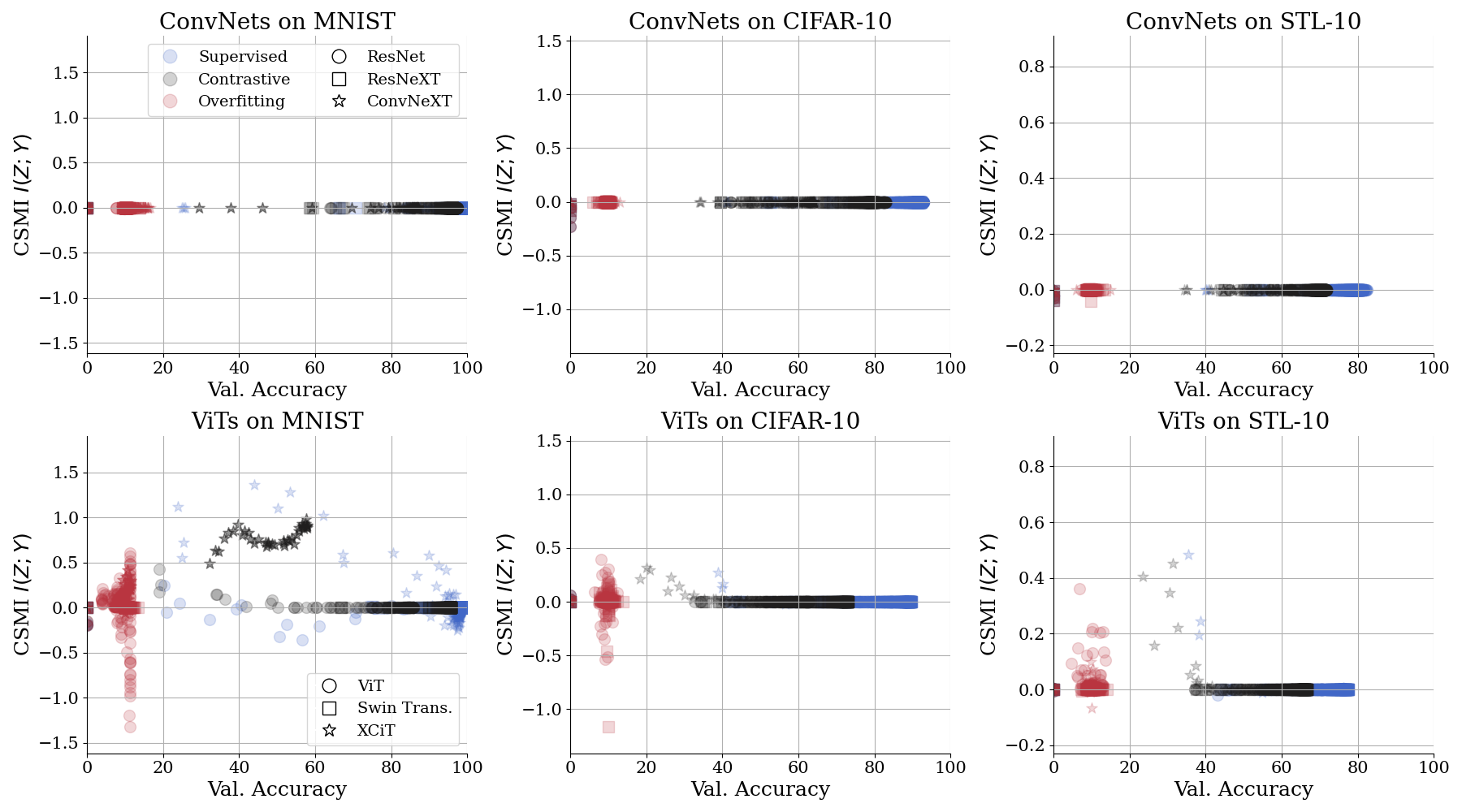}
    \caption{\textbf{Classic Shannon mutual information version for Figure~\ref{fig:main_results_DSMI_output}.}}
    \label{fig:main_results_CSMI_output}
\end{figure*}

\begin{figure*}[!htb]
    \centering
    \includegraphics[width=0.7\textwidth]{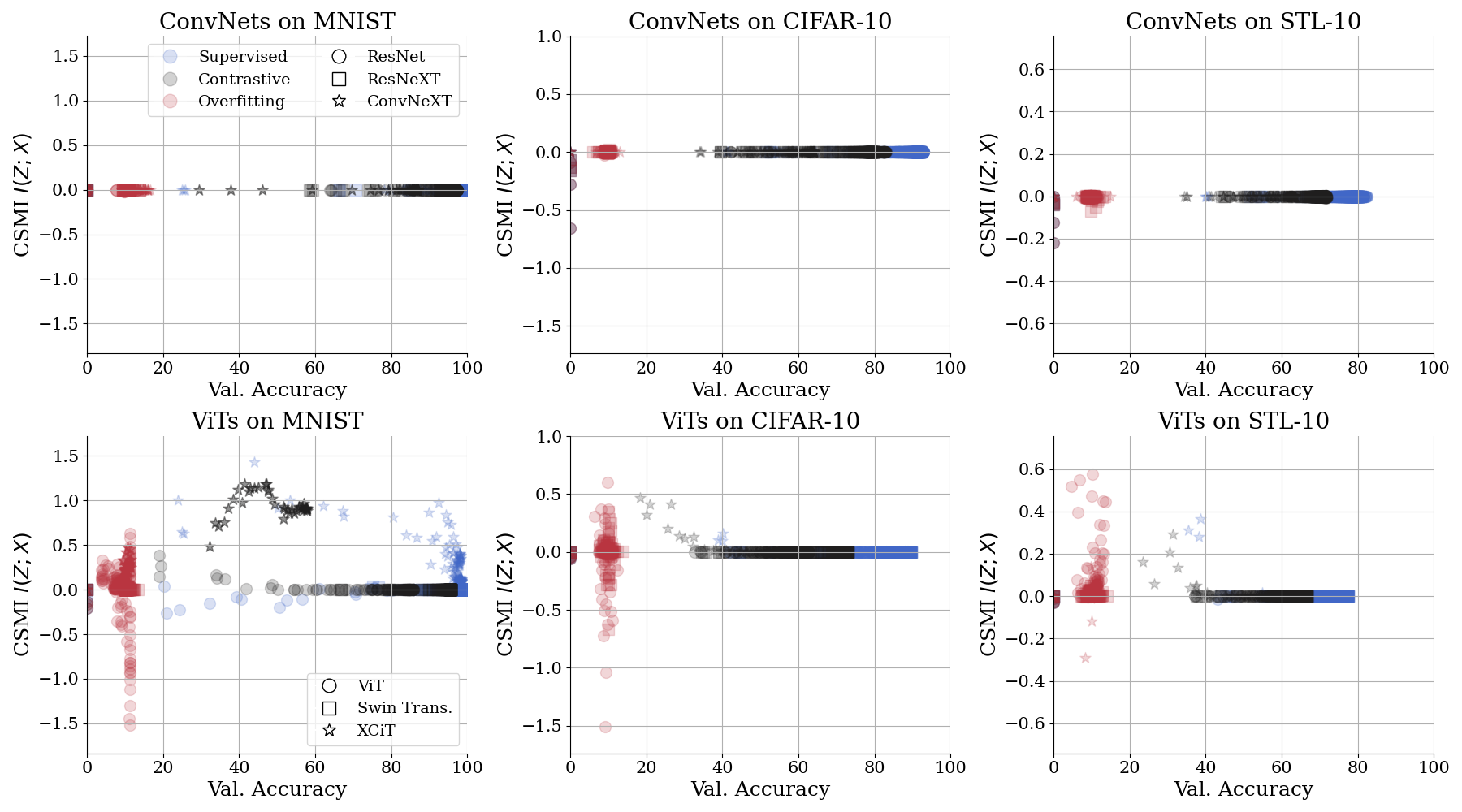}
    \caption{\textbf{Classic Shannon mutual information version for Figure~\ref{fig:main_results_DSMI_input}.}}
    \label{fig:main_results_CSMI_input}
\end{figure*}

This phenomenon can be seen in both toy data (Figures~\ref{fig:main_results_DSE} and \ref{fig:main_results_DSMI_output}) and real data (Figures~\ref{fig:main_results_CSE}, \ref{fig:main_results_CSMI_output} and \ref{fig:main_results_CSMI_input}).

\clearpage
\newpage

\section{Propositions with Proofs on DSE and DSMI}
\label{supp:proposition_proofs}
The following propositions establish some bounds on the minimal and maximal values of Diffusion Spectral Entropy~(DSE). In addition, they provide insight on the definition of Diffusion Spectral Mutual Information~(DSMI). Note that taking $t\rightarrow \infty$ allows us to talk about the major structures in the dataset. 



\textbf{Proposition~\ref{prop:minimal_entropy}}
$S_D$ achieves a minimal entropy of $0$ when the diffusion operator defines an ergodic Markov chain, and is in steady state (as $t \rightarrow \infty$).

\begin{proof}
The eigenvalues of a finite ergodic Markov chain have the form $1 = |\lambda_1| > |\lambda_2| \geq \cdots \geq |\lambda_N| \geq 0$, we see that $\lambda_i^t = 0$ as $t \rightarrow \infty \hspace{4pt} \forall i > 1$. Thus, the resultant entropy is $1 \log(1) + \sum_{i>1} 0 = 0$, proving the proposition.
\end{proof}

\noindent\textbf{Proposition~\ref{prop:cluster_entropy}}
As $t \rightarrow \infty$, $S_D(\mP_X,t)$ on data with $k$ well-separated, connected clusters is $\log k$. 

\begin{proof}
If the data has $k$ well-separated clusters, then $\mP_X$ has eigenvalues of the form $1 = |\lambda_1| = |\lambda_2| = \cdots = |\lambda_k| > |\lambda_{k+1}| \ge \cdots \ge |\lambda_N| \ge 0$. In other words, the multiplicity of $1$ eigenvalues of $\mP_X$ corresponds to the number of connected components in the underlying graph, which here is $k$, all other eigenvalues are strictly less than $1$ and greater than or equal to $0$. Therefore, as $t\rightarrow \infty$ only these eigenvalues remain and the resultant DSE is $\sum_k \frac{1}{k} \log (k) = \log (k)$ completing the proof.
\end{proof}

\begin{lemma}\label{lemma:semidef}
    Let $K$ be the anisotropic kernel matrix, where for data points $x_i$ and $x_j$, $K_{ij} = \frac{G(x_i, x_j)}{\|G(x_i, \cdot)\|_1^\alpha\|G(x_j, \cdot)\|_1^\alpha}$, and $G(x_i, x_j) = \exp(-\|x_i - x_j\|^2_2/\sigma)$. Then, the anisotropic kernel matrix $K$ is a positive semidefinite matrix.
\end{lemma}
\begin{proof}
    Note that $K = D^{-\alpha} G D^{-\alpha}$, where $G$ is the Gaussian kernel matrix and $D_{ii} = \sum_{j=1}^n G_{ij}$ for all $i \in [n]$, and $D_{ij} = 0$ for $i \neq j$. It is well known that the Gaussian kernel matrix $G$ is positive semidefinite, so $x^TGx \geq 0$ $\forall x \in \mathbb{R}^n \backslash \{ 0\}$. Let $\tilde{x} = D^{-\alpha}x \in \mathbb{R}^n \backslash \{ 0\}$, and $\tilde{x}^TK\tilde{x} = (x^TD^{-\alpha}) G (D^{-\alpha}x) \geq 0$ $\forall \tilde{x} \in \mathbb{R}^n \backslash \{ 0\}$ since $\tilde{x}$ is a linear reparameterization of $x$.
\end{proof}

\begin{lemma}\label{lemma:telescope}
    Given $n$ data points, let
    \begin{align*}
        \phi_m(\lambda_1, \cdots, \lambda_m) =& -(n - m + 1)\frac{\lambda_m}{(n - m + 1)\lambda_m + \sum_{j=1}^{m-1} \lambda_j}\log\left(\frac{\lambda_m}{(n - m + 1)\lambda_m + \sum_{j=1}^{m-1} \lambda_j}\right)
        \\& - \sum_{i = 1}^{m-1}\frac{\lambda_i}{(n - m + 1)\lambda_m + \sum_{j=1}^{m-1} \lambda_j}\log\left(\frac{\lambda_i}{(n - m + 1)\lambda_m + \sum_{j=1}^{m-1} \lambda_j}\right)
    \end{align*}
    be a function of the eigenvalues of the anisotropic diffusion matrix $K$ for $m \in \{2, \cdots, n\}$, where the eigenvalues are ordered such that $\lambda_1 \geq \lambda_2 \geq \cdots \geq \lambda_n$. For all $m \in \{2, n-1\}$, $\phi_{m}(\lambda_1, \cdots, \lambda_{m}) \geq \phi_{m+1}(\lambda_1, \cdots, \lambda_{m}, \lambda_{m+1})$.
\end{lemma}

\begin{proof}
    By Lemma \ref{lemma:semidef}, $K$ has all non-negative eigenvalues. Staying in convention with DSE, let $0\log\left(\frac{0}{0}\right) = 0$.

    First, we observe the following:

    $$\phi_{m}(\lambda_1, \cdots, \lambda_m) = \phi_{m+1}(\lambda_1, \cdots, \lambda_m, \lambda_m)$$

    By the Fundamental Theorem of Calculus,

    $$\int_{\lambda_{m+1}}^{\lambda_{m}}\frac{\partial\phi_{m+1}}{\partial\lambda_{m+1}}d\lambda_{m+1} = \phi_{m} - \phi_{m+1}$$

    Thus, it suffices to show that on $[\lambda_{m}, \lambda_{m+1}]$, $\int_{\lambda_{m+1}}^{\lambda_{m}}\frac{\partial\phi_{m+1}}{\partial\lambda_{m+1}}d\lambda_{m+1} \geq 0$. In pursuit of tidiness, let $\phi_{m+1} = f + g$, where $f(\lambda_1, \cdots, \lambda_{m+1}) = (n - m)\frac{\lambda_{m+1}}{(n - m)\lambda_{m+1} + \sum_{j=1}^{m} \lambda_j}\log\left(\frac{\lambda_{m+1}}{(n - m)\lambda_{m+1} + \sum_{j=1}^{m} \lambda_j}\right)$, and $g$ represents the summation term.

    For $f$,
    \begin{equation*}
        \begin{split}
            \frac{\partial f}{\partial\lambda_{m+1}} &= -(n-m)\left(\log\left(\frac{\lambda_{m+1}}{(n - m)\lambda_{m+1} + \sum_{j=1}^{m} \lambda_j}\right) + 1\right)\\&\hspace{10pt}\left[\frac{-(n-m)\lambda_{m+1}}{\left((n-m)\lambda_{m+1}+\sum_{j=1}^{m}\lambda_j\right)^2}+\frac{1}{(n-m)\lambda_{m+1}+\sum_{j=1}^{m}\lambda_j}\right] \\
            &= -(n-m)\left(\log\left(\frac{\lambda_{m+1}}{(n - m)\lambda_{m+1} + \sum_{j=1}^{m} \lambda_j}\right) + 1\right)\left[\frac{-(n-m)\lambda_{m+1}+(n-m)\lambda_{m+1}+\sum_{j=1}^{m}\lambda_j}{\left((n-m)\lambda_{m+1}+\sum_{j=1}^{m}\lambda_j\right)^2}\right] \\
            &= -(n-m)\left(\log\left(\frac{\lambda_{m+1}}{(n - m)\lambda_{m+1} + \sum_{j=1}^{m} \lambda_j}\right) + 1\right)\left[\frac{\sum_{j=1}^{m}\lambda_j}{\left((n-m)\lambda_{m+1}+\sum_{j=1}^{m}\lambda_j\right)^2}\right]
        \end{split}
    \end{equation*}

    For $g$,
    \begin{equation*}
        \begin{split}
            \frac{\partial g}{\partial\lambda_{m+1}} &= -\sum_{i=1}^m\left(\log\left(\frac{\lambda_{i}}{(n - m)\lambda_{m+1} + \sum_{j=1}^{m} \lambda_j}\right) + 1\right)\left[\frac{-(n-m)\lambda_i}{\left((n-m)\lambda_{m+1}+\sum_{j=1}^{m}\lambda_j\right)^2}\right] \\
            &= \sum_{i=1}^m\left(\log\left(\frac{\lambda_{i}}{(n - m)\lambda_{m+1} + \sum_{j=1}^{m} \lambda_j}\right) + 1\right)\left[\frac{(n-m)\lambda_i}{\left((n-m)\lambda_{m+1}+\sum_{j=1}^{m}\lambda_j\right)^2}\right]
        \end{split}
    \end{equation*}

    Together, letting $S = (n - m)\lambda_{m+1} + \sum_{j=1}^{m} \lambda_j$,
    \begin{equation*}
        \begin{split}
            \frac{\partial\phi_{m+1}}{\partial\lambda_{m+1}} &= \frac{\partial g}{\partial\lambda_{m+1}}+\frac{\partial f}{\partial\lambda_{m+1}} \\
            &= \frac{1}{S^2}\left((n-m)\sum_{i=1}^m\lambda_i\log\left(\frac{\lambda_i}{S}\right)+(n-m)\sum_{i=1}^m\lambda_i -(n-m)\sum_{i=1}^m\lambda_{i}\log\left(\frac{\lambda_{m+1}}{S}\right)-(n-m)\sum_{i=1}^m\lambda_i\right) \\
            &= \frac{1}{S^2}\left((n-m)\sum_{i=1}^m\lambda_i\log\left(\frac{\lambda_i}{S}\right) -(n-m)\sum_{i=1}^m\lambda_{i}\log\left(\frac{\lambda_{m+1}}{S}\right)\right) \\
            &= \frac{n-m}{S^2}\sum_{i=1}^m\lambda_i\log\left(\frac{\lambda_i}{\lambda_{m+1}}\right)
        \end{split}
    \end{equation*}

    We conclude by observing that $\lambda_{m+1}$ is the smallest eigenvector, so every log term must be non-negative.
    
\end{proof}

Here, note that $\phi_n = S_D(\mathbf{P}_X, t = 1)$.

\begin{lemma}\label{lemma:concave}
    Define $\phi$ as in Lemma \ref{lemma:telescope}. Then, the function $\phi_2(\lambda_1, \lambda_2)$ is concave for the anisotropic diffusion matrix.
\end{lemma}

\begin{proof}
    Since $\lambda_1 \equiv 1$ by construction, it suffices to show that $\frac{\partial^2\phi_2}{\partial\lambda_2^2}(1, \lambda_2) \leq 0$ for all $\lambda_2 \leq 1$.

    \begin{equation*}
        \begin{split}
            \frac{\partial\phi_2}{\partial\lambda_2} &= -\left(\log\left(\frac{1}{1+(n-1)\lambda_2}\right) + 1\right)\frac{-1(n-1)}{(1 + (n-1)\lambda_2)^2} \\
            &-(n-1)\left(\log\left(\frac{\lambda_2}{1+(n-1)\lambda_2}\right) + 1\right)\left(\frac{-\lambda_2(n-1)}{(1+(n-1)\lambda_2)^2} + \frac{1}{1+(n-1)\lambda_2}\right) \\
            &= \frac{(n-1)}{(1 + (n-1)\lambda_2)^2}\left(\log\left(\frac{1}{1+(n-1)\lambda_2}\right) + 1\right) \\
            &+ \frac{(n-1)}{(1 + (n-1)\lambda_2)^2}\left(\log\left(\frac{\lambda_2}{1+(n-1)\lambda_2}\right) + 1\right)\left(\lambda_2(n-1) - (1 + (n-1)\lambda_2\right) \\
            &= \frac{(n-1)}{(1 + (n-1)\lambda_2)^2}\left(\log\left(\frac{1}{1+(n-1)\lambda_2}\right) + 1 - \log\left(\frac{\lambda_2}{1+(n-1)\lambda_2}\right) - 1\right) \\
            &= \frac{(n-1)}{(1 + (n-1)\lambda_2)^2}\left(\log\left(\frac{1}{\lambda_2}\right)\right) \\
            &= -\frac{(n-1)}{(1 + (n-1)\lambda_2)^2}\log\left(\lambda_2\right)
        \end{split}
    \end{equation*}

    \begin{equation*}
        \begin{split}
            \frac{\partial^2\phi_2}{\partial\lambda^2_2} &= -(-2)(n-1)\frac{(n-1)}{(1 + (n-1)\lambda_2)^3}\log\left(\lambda_2\right) -\frac{(n-1)}{\lambda_2(1 + (n-1)\lambda_2)^2} \\
            &= 2(n-1)\frac{(n-1)}{(1 + (n-1)\lambda_2)^3}\log\left(\lambda_2\right) -\frac{(n-1)(1 + (n-1)\lambda_2)}{\lambda_2(1 + (n-1)\lambda_2)^3} \\
            &= \frac{n-1}{(1 + (n-1)\lambda_2)^3}\left(2(n-1)\log\left(\lambda_2\right) -\frac{(1 + (n-1)\lambda_2)}{\lambda_2}\right) \\
            &= \frac{n-1}{(1 + (n-1)\lambda_2)^3}\left(2(n-1)\log\left(\lambda_2\right) - \left(\frac{1}{\lambda_2} + (n-1)\right)\right) \\
        \end{split}
    \end{equation*}

    Concavity requires $\frac{1}{\lambda_2} + (n - 1) \geq 2(n-1)\log(\lambda_2)$. Here, we observe that $\lambda_2 \leq 1$, and $\frac{1}{\lambda_2} + (n - 1) > 0 \geq 2(n-1)\log(\lambda_2)$ completes the proof.
\end{proof}

\noindent\textbf{Proposition~\ref{prop:random-gaussian}}
    Let $X \in \mathbb{R}^{n\times d}$ be a dataset of $n$ independent and identically distributed multivariate Gaussian vectors in $\mathbb{R}^d$, where $x_i \sim \mathcal{N}(0, I_d)$. Then, using $K$ as defined in Eqn~\ref{eqn:diffusion_kernel} with $\alpha = 1/2$,

    \begin{equation}
    \begin{aligned}
    &\mathbb{E}[S_D(\mathbf{P}_X, t = 1)] \\\lessapprox& \log(\frac{n}{1-\beta}) - \left(\frac{1}{n} + \left(\frac{n-1}{n}\right)\beta\right) \log\left(1+\frac{\beta n}{1 - \beta}\right)\\
    &\mathrm{where} \hspace{4pt} \beta = \left(1 + \frac{4}{\sigma}\right)^{-\frac{d}{2}}
    \end{aligned}
    \end{equation}

\begin{proof}
    We prove the above theorem in four steps:
    \begin{enumerate}[noitemsep,nolistsep]
        \item Expectation is linear.
        \item One can approximate the expected random anisotropic kernel matrix.
        \item Based on this, one can derive the expected eigenvalues of the expected matrix.
        \item Use Jensen's Inequality with the bounded DSE.
    \end{enumerate}

    First, eigenvalues $\lambda$ are defined such that for matrix $K\in \mathbb{R}^{n\times n}$, $Kx = \lambda x$ for all vectors $x \in \mathbb{R}^n$. By the linearity of expectation, $\mathbb{E}[K]x = \mathbb{E}[\lambda] x$. Thus, computing $\mathbb{E}[K]$ leads to a solution for $\mathbb{E}[\lambda]$, which can be used to upper bound the DSE. 

    Second, $\mathbb{E}[K] = \mathbb{E}[D^{-1/2}GD^{-1/2}]$. Here, we approximate the expectation by $\mathbb{E}[K] \approx (D\mid\mathbb{E}[G])^{-1/2}\mathbb{E}[G](D\mid\mathbb{E}[G])^{-1/2}$, as confirmed empirically. Computations based on $\mathbb{E}[G]$ are much more approachable, given the nature of the problem. 
    
    For diagonal entries of $G$, $\mathbb{E}[G_{ii}] = \mathbb{E}\left[\exp\left(-\|x_i - x_i\|_2^2 / \sigma\right)\right] = 1$. For off-diagonal entries of $G$, we proceed inductively. First, $x_i - x_j \sim \mathcal{N}(0, 2I_d)$. Let $z_{ij} \equiv \|x_i - x_j\|^2_2$. Since $\frac{z_{ij}}{2} \sim \chi^2_d$, one can verify that the chi-squared distribution moment generating function is $\mathbb{E}[e^{tX}] = (1-2t)^{-d/2}$ for $t < 1/2$. Thus, $\mathbb{E}\left[\exp\left(-\|x_i - x_j\|_2^2 / \sigma\right)\right] = \mathbb{E}[e^{(-1/\sigma)2z_{ij}}] = (1 + \frac{4}{\sigma})^{-d/2}$.

    For $\mathbb{E}[K]$,
    \begin{align*}
        \mathbb{E}[K_{ii}] &\approx \frac{1}{1+(n-1)(1 + \frac{4}{\sigma})^{-d/2}} \\
        \mathbb{E}[K_{ij}] &\approx \frac{(1 + \frac{4}{\sigma})^{-d/2}}{1+(n-1)(1 + \frac{4}{\sigma})^{-d/2}}\quad\forall i \neq j
    \end{align*}

    Third, observe that our approximate $\mathbb{E}[K]$ may be decomposed as follows:
    \begin{align*}
        \mathbb{E}[K] \approx \mathbbm{1}\mathbbm{1}^T\left(\frac{(1 + \frac{4}{\sigma})^{-d/2}}{1+(n-1)(1 + \frac{4}{\sigma})^{-d/2}}\right) + I_n\left(\frac{1-(1 + \frac{4}{\sigma})^{-d/2}}{1+(n-1)(1 + \frac{4}{\sigma})^{-d/2}}\right)
    \end{align*}

    One easily verifies that the first matrix has eigenvalues $\lambda_1 = n\left(\frac{(1 + \frac{4}{\sigma})^{-d/2}}{1+(n-1)(1 + \frac{4}{\sigma})^{-d/2}}\right)$ and $\lambda_2 = \cdots = \lambda_n = 0$, and the second matrix has eigenvalues $\lambda_1 = \cdots = \lambda_n = \frac{1-(1 + \frac{4}{\sigma})^{-d/2}}{1+(n-1)(1 + \frac{4}{\sigma})^{-d/2}}$. By construction, the eigenvalues are additive, so $\mathbb{E}[K]$ has $\mathbb{E}[\lambda_1] = 1$, and $\mathbb{E}[\lambda_2] = \cdots = \mathbb{E}[\lambda_n] = \frac{1-(1 + \frac{4}{\sigma})^{-d/2}}{1+(n-1)(1 + \frac{4}{\sigma})^{-d/2}}$.

    Fourth, we refer to Lemma \ref{lemma:telescope} in claiming that $\phi_2(\lambda_1, \lambda_2) \geq \phi_n(\lambda_1, \lambda_2, \cdots, \lambda_n) = S_D(\mathbf{P}_X, t = 1)$. By the concavity established in Lemma \ref{lemma:concave}, Jensen's Inequality implies that $\mathbb{E}[\phi_2(\lambda_1, \lambda_2)] \leq \phi_2(\mathbb{E}[\lambda_1], \mathbb{E}[\lambda_2])$. Simply put,

    \begin{align*}
    \mathbb{E}[S_D(\mathbf{P}_X, t = 1)] \leq \phi_2(\mathbb{E}[\lambda_1], \mathbb{E}[\lambda_2])        
    \end{align*}

    \clearpage
    Finally, letting $\beta = (1 + \frac{4}{\sigma})^{-d/2}$,
    \begin{equation*}
        \begin{split}
            \phi_2(\mathbb{E}[\lambda_1], \mathbb{E}[\lambda_2]) &= -\frac{(n - 1)\mathbb{E}[\lambda_2]}{(n - 1)\mathbb{E}[\lambda_2] + \mathbb{E}[\lambda_1]}\log\left(\frac{\mathbb{E}[\lambda_2]}{(n - 1)\mathbb{E}[\lambda_2] + \mathbb{E}[\lambda_1]}\right) \\
            &\hspace{12pt} - \frac{\mathbb{E}[\lambda_1]}{(n - 1)\mathbb{E}[\lambda_2] +  \mathbb{E}[\lambda_1]}\log\left(\frac{\mathbb{E}[\lambda_1]}{(n - 1)\mathbb{E}[\lambda_2] +  \mathbb{E}[\lambda_1]}\right) \\
            &= -\frac{(n - 1)\frac{1-\beta}{1+(n-1)\beta}}{(n - 1)\frac{1-\beta}{1+(n-1)\beta} + 1}\log\left(\frac{\frac{1-\beta}{1+(n-1)\beta}}{(n - 1)\frac{1-\beta}{1+(n-1)\beta} + 1}\right) \\
            &\hspace{12pt} - \frac{1}{(n - 1)\frac{1-\beta}{1+(n-1)\beta} + 1}\log\left(\frac{1}{(n - 1)\frac{1-\beta}{1+(n-1)\beta} + 1}\right) \\
            &= -(n - 1)\frac{(1-\beta)}{(n - 1)(1-\beta) + 1+(n-1)\beta}\log\left(\frac{(1-\beta)}{(n - 1)(1-\beta) + 1+(n-1)\beta}\right) \\
            &\hspace{12pt} - \frac{1+(n-1)\beta}{(n - 1)(1-\beta) + 1+(n-1)\beta}\log\left(\frac{1+(n-1)\beta}{(n - 1)(1-\beta) + 1+(n-1)\beta}\right) \\
            &= -(n - 1)\frac{(1-\beta)}{(n - 1) + 1}\log\left(\frac{(1-\beta)}{(n - 1)+ 1}\right)- \frac{1+(n-1)\beta}{(n - 1)+ 1}\log\left(\frac{1+(n-1)\beta}{(n - 1) + 1}\right) \\
            &= -(n - 1)\frac{(1-\beta)}{n}\log\left(\frac{(1-\beta)}{n}\right)- \frac{1+(n-1)\beta}{n}\log\left(\frac{1+(n-1)\beta}{n}\right) \\
            &= \frac{n - 1 - n\beta + \beta}{n}\log\left(\frac{n}{1-\beta}\right)+ \frac{1+n\beta-\beta}{n}\log\left(\frac{n}{1+(n-1)\beta}\right) \\
            &= \frac{n}{n}\log\left(\frac{n}{1-\beta}\right)+ \frac{1+n\beta-\beta}{n}\left(\log\left(\frac{n}{1+(n-1)\beta}\right) - \log\left(\frac{n}{1-\beta}\right)\right) \\
            &= \log\left(\frac{n}{1-\beta}\right)+ \frac{1+n\beta-\beta}{n}\left(\log\left(\frac{1-\beta}{1+(n-1)\beta}\right)\right) \\
            &= \log\left(\frac{n}{1-\beta}\right)+ \left(\frac{1}{n} +\frac{n-1}{n}\beta\right)\log\left(\frac{1-\beta}{1+(n-1)\beta}\right) \\
            &= \log\left(\frac{n}{1-\beta}\right) - \left(\frac{1}{n} +\frac{n-1}{n}\beta\right)\log\left(\frac{1-\beta+n\beta}{1-\beta}\right) \\
            &= \log\left(\frac{n}{1-\beta}\right) - \left(\frac{1}{n} +\frac{n-1}{n}\beta\right)\log\left(1+\frac{n\beta}{1-\beta}\right) \\
        \end{split}
    \end{equation*}
    
\end{proof}

\begin{cor}
\label{cor:maximal_entropy}
$S_D$ achieves maximal entropy in a matrix where each point only transitions to itself, and the entropy here will be $\log (n)$.
\end{cor}

\begin{proof}
In this case the transition matrix corresponds to the identity matrix, hence, each of the $n$ eigenvalues is $1$. Thus the diffusion spectral entropy is the uniform distribution on $n$ states $-\sum_{i}^n  (1/n) \log(1/n)= \log(n)$, which maximizes the entropy.  
\end{proof}

\begin{lemma}\label{lemma:blocks}
Let $f$ be a probability distribution, and let $X \in \mathbb{R}^{kn \times d}$ be a dataset of $n \in \mathbb{N}$ independent and identically distributed $d$-dimensional points $x_i \sim f$. Let these points lie in $k \in \mathbb{N}$ distinct clusters, where the diffusion probability between points of different clusters is arbitrarily small. Let $Y \in \mathbb{R}^{n \times d}$ be a dataset of $n$ i.i.d. points $y_i \sim f$ in one cluster. Then, the expected DSE $\mathbb{E}[S_D(\mathbf{P}_X, t)]$ may be related to the expected DSE $\mathbb{E}[S_D(\mathbf{P}_Y, t)]$ by the following equality:

$$\mathbb{E}\left[S_D(\mathbf{P}_X, t)\right] = \mathbb{E}\left[S_D(\mathbf{P}_Y, t)\right] + \log(k)$$
\end{lemma}

\begin{proof}
Let $K \in \mathbb{R}^{n\times n}$ be the anisotropic kernel matrix for $Y$, and let $\lambda_1, \cdots, \lambda_n$ denote its ordered eigenvalues. Let $\tilde{K} \in \mathbb{R}^{kn\times kn}$ be the anisotropic kernel matrix for $X$, where $\tilde{\lambda}_1, \cdots, \tilde{\lambda}_{kn}$ are its ordered eigenvalues. Note that as a block diagonal matrix, $\tilde{\lambda}_{ki+1} = \cdots \tilde{\lambda}_{ki+k}$ in law for $i \in [0, n-1]$. Since each block is composed of $n$ data points distributed as $f$, $\lambda_{i+1} = \tilde{\lambda}_{ki+1}$ in law. Thus, we can reformulate $\mathbb{E}[S_D(\mathbf{P}_X, t)]$ as follows:

    \begin{equation*}
        \begin{split}
            \mathbb{E}[S_D(\mathbf{P}_X, t)] &= \mathbb{E}\left[-\sum_{i=1}^{kn}\left(\frac{|\tilde{\lambda}^t_i|}{\sum_{i=1}^{kn}|\tilde{\lambda}^t_j|}\right) \log\left(\frac{|\tilde{\lambda}^t_i|}{\sum_{i=1}^{kn}|\tilde{\lambda}^t_j|}\right)\right]\\
            &= \mathbb{E}\left[-k\cdot\sum_{i=1}^{n}\left(\frac{|\lambda^t_i|}{k\cdot\sum_{i=1}^{n}|\lambda^t_j|}\right) \log\left(\frac{|\lambda^t_i|}{k\cdot\sum_{i=1}^{n}|\lambda^t_j|}\right)\right]\\
            &= \mathbb{E}\left[-\frac{k}{k}\cdot\sum_{i=1}^{n}\left(\frac{|\lambda^t_i|}{\sum_{i=1}^{n}|\lambda^t_j|}\right) \log\left(\frac{|\lambda^t_i|}{k\cdot\sum_{i=1}^{n}|\lambda^t_j|}\right)\right]\\
            &= \mathbb{E}\left[-\sum_{i=1}^{n}\left(\frac{|\lambda^t_i|}{\sum_{i=1}^{n}|\lambda^t_j|}\right) \left(\log\left(\frac{|\lambda^t_i|}{\sum_{i=1}^{n}|\lambda^t_j|}\right) - \log(k)\right)\right]\\
            &= \mathbb{E}\left[-\sum_{i=1}^{n}\left(\frac{|\lambda^t_i|}{\sum_{i=1}^{n}|\lambda^t_j|}\right) \log\left(\frac{|\lambda^t_i|}{\sum_{i=1}^{n}|\lambda^t_j|}\right)\right] - \log(k)\cdot\mathbb{E}\left[-\sum_{i=1}^{n}\left(\frac{|\lambda^t_i|}{\sum_{i=1}^{n}|\lambda^t_j|}\right) \right]\\
            &= \mathbb{E}[S_D(\mathbf{P}_Y, t)] + \log(k)\\
        \end{split}
    \end{equation*}
\end{proof}

\noindent\textbf{Proposition~\ref{prop:one-to-k}}
Take $n$ to be arbitrarily large. Let $X \in \mathbb{R}^{n\times d}$ be a matrix of i.i.d. random values $x_{ij} \sim f$. Let $Y \in \mathbb{R}^{n\times d}$ be a matrix of i.i.d. random values $y_{ij} \sim f$, but in $k \in \mathbb{N}$ distinct clusters such that when the anisotropic probability matrix is computed for $\alpha = 1/2$, the probability of diffusion between points of different clusters is arbitrarily small. Then, using $\beta$ as defined in Proposition \ref{prop:random-gaussian}, the approximate upper bound on DSE increases by $\beta\log(k)$.

\begin{proof}
    We take $n$ to be large such that the upper bound in Prop. \ref{prop:random-gaussian} approaches $\log\left(\frac{n}{1-\beta}\right) - \beta\log\left(\frac{\beta n}{1-\beta}\right)$, for $\mathbb{E}[S_D(\mathbf{P}_X, t = 1)]$. Let $Z \in \mathbb{R}^{n/k\times d}$ also contain i.i.d. random entries according to $f$. For $\mathbb{E}[S_D(\mathbf{P}_Z, t = 1)]$, the aforementioned upper bound evaluates to $\log\left(\frac{n}{1-\beta}\right) - \beta\log\left(\frac{\beta n}{1-\beta}\right) - (1-\beta)\log(k)$ following simple manipulations.

    By Lemma~\ref{lemma:blocks}, $\mathbb{E}[S_D(\mathbf{P}_Y, t = 1)] = \mathbb{E}[S_D(\mathbf{P}_Z, t = 1)] + \log(k)$.

    \begin{equation*}
        \begin{split}
            \mathbb{E}[S_D(\mathbf{P}_Y, t = 1)] &= \mathbb{E}[S_D(\mathbf{P}_Z, t = 1)] + \log(k) \\
            &\leq \log\left(\frac{n}{1-\beta}\right) - \beta\log\left(\frac{\beta n}{1-\beta}\right) - (1-\beta)\log(k) + \log(k) \\
            &\leq \log\left(\frac{n}{1-\beta}\right) - \beta\log\left(\frac{\beta n}{1-\beta}\right) + \beta\log(k) \\
        \end{split}
    \end{equation*}
\end{proof}

\noindent\textbf{Proposition~\ref{prop:cluster_mi}}
As $t \rightarrow \infty$ in a hidden layer $X$ with $k$ well-separated clusters and class labels $Y$ perfectly indicating clusters, we will have $I_D(X;Y)=\log(k)$. 

\begin{proof}
Based on proposition~\ref{prop:cluster_entropy}, the entropy of $k$ well-separated clustered data with $t\rightarrow \infty$ is $\log(k)$, thus $S_D(\mathbf{P}_x,t)=log(k)$. However, for each label $y$ $S_D(\mathbf{P}_x, t)= \log(1)=0$. Thus $I_D(X;Y)= \log k$.
\end{proof}

\newpage
\section{Full Caption of the Toy Data Figure}
\label{sec:full_caption_toy_data_entropy_MI}

Here we show Figure~\ref{fig:toy_data_entropy_MI} with the full caption.

\begin{figure*}[!bh]
    \centering
    \includegraphics[width=\textwidth]{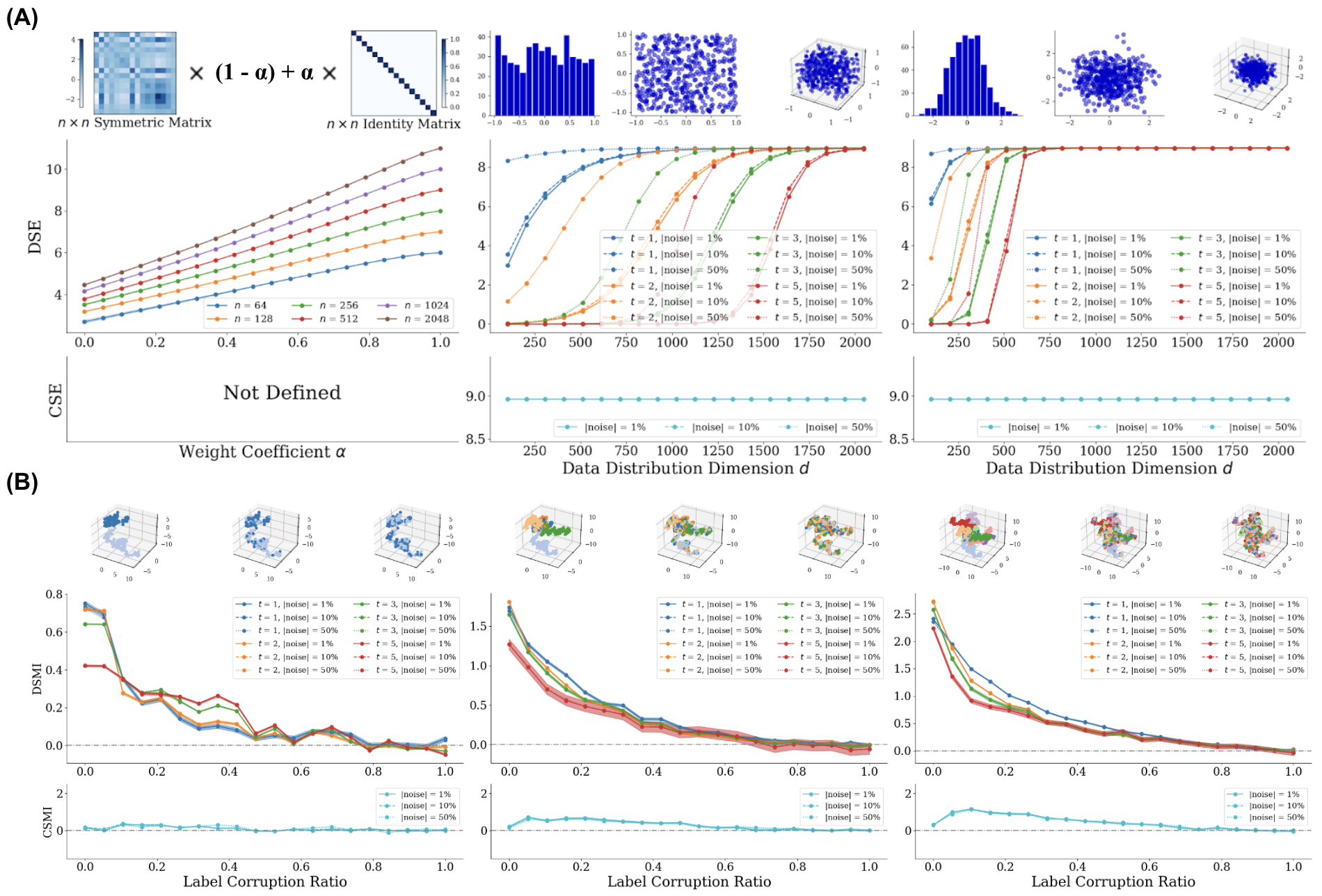}
    \caption{\textbf{Diffusion Spectral Entropy~(DSE) and Diffusion Spectral Mutual Information~(DSMI) on toy data.}\\ \textbf{(A)} \textit{DSE increases as intrinsic dimension grows, while classic Shannon entropy~(CSE) saturates to $\log(n)$ due to curse of dimensionality.} \textbf{Left}: Weighted sum of a random $n \times n$ symmetric positive definite matrix (to simulate a diffusion matrix) and the $n \times n$ identity matrix. In theory, the identity matrix shall have the highest entropy at each respective $n$. \textbf{Mid}: $d$-dimensional $U[-1, 1]$ inside a $2048$ dimensional space. \textbf{Right}: $d$-dimensional $\mathcal{N}(0, I)$ inside a $2048$ dimensional space. Shaded areas indicate the standard deviation from 5 independent runs. For the latter two distributions, additive noise is injected into the coordinates and schematics for $d=\{1, 2, 3\}$ are illustrated on top. The number of data points for the simulation is 500. Note that in the latter two cases for DSE we compute the diffusion matrix of the data manifold prior to entropy evaluation, whereas in the first case we skip that step because the matrix is already provided. In the latter two cases, CSE saturates to $\log(n)=\log_2(500)=8.966$. \\
    \textbf{(B)} \textit{When two random variables are dependent, DSMI negatively correlates with the level of data corruption, while classic Shannon mutual information~(CSMI) does not capture this trend.} DSMI $I_D(Z; Y)$ and CSMI are computed on synthetic, 20-dimensional trees with \{2, 5, 10\} branches (\textbf{Left, Mid, Right}). The x-axis represents the level of label corruption, ranging from completely clean label (left) to fully corrupted label (right), with 3 dimensional schematics demonstrating corruption ratios \{0.0, 0.5, 1.0\} displayed on top. At full corruption, $I_D(Z; Y)$ converges to zero, as the embedding vectors do not contain information on the labels.}
\end{figure*}


\newpage
\section{Experimental Details}
\label{supp:experiment_details}

\subsection{Three conditions for neural network training process}
All experiments were run on a single NVIDIA A100 GPU.

\subsubsection{Supervised learning}
We trained the 6 networks (3 ConvNets and 3 vision transformers as mentioned in Section~\ref{sec:real_exp}) end-to-end. The ConvNets included ResNet~\citeSupp{ResNet}, ResNeXT~\citeSupp{ResNeXT}, and ConvNeXT~\citeSupp{ConvNeXT}. The vision transformers included the original Vision Transformer~(ViT)~\citeSupp{ViT}, Swin Transformer~\citeSupp{SwinT}, and Cross-Covariance Image Transformer~(XCiT)~\citeSupp{XCiT}.

We used an AdamW optimizer~\citeSupp{AdamW} with an initial learning rate of 1e-5 for MNIST~\citeSupp{MNIST} or 1e-4 for CIFAR-10~\citeSupp{CIFAR10} and STL-10~\citeSupp{STL10}. The learning rate is modulated by a Cosine Annealing Scheduler~\citeSupp{CosineAnnealing} with a linear warm-up for the first 10 epochs. The total duration of training is 200 epochs. At the end of each epoch, we pass the entire validation set through the model and collect the $\mathbb{R}^{D}$ representation vectors as output from the penultimate layer for further analysis.

\subsubsection{Contrastive learning}
For the contrastive learning experiments, we followed the SimCLR~\citeSupp{SimCLR} paradigm: we create two augmentations of the same image and ask the model to embed them closer on the embedding manifold, while encouraging greater separation between them and the other images in the same batch. The standard training procedure of contrastive learning first trains the backbone, with the classifier (the final, linear layer) detached, for some epochs, and then performs a linear probing or fine-tuning. In either case, a linear classifier layer is attached and trained for some more epochs. The weights of the backbone are frozen in the former case, versus learnable in the latter.

Since we need to assess how well the model is learning at each epoch, we instead perform linear probing by the end of each epoch. Specifically, we freeze the backbone weights, attach a reinitialized linear classifier layer, and train the classifier for 10 epochs with the training set. Then, we record the end-to-end validation accuracy and collect the embedding vectors on the validation set, similar to the supervised learning case. Finally, we unfreeze the backbone weights for the next epoch of SimCLR training. For fair comparison, the training details are otherwise the same as in the supervised learning case, including the learning rate and scheduling.

\subsubsection{Purposeful overfitting}
In purposeful overfitting, we train the model in the same way as in the supervised learning case, except that the data labels are randomly permuted. In this way, the models are forced to learn nonsense labels. To better overfit, we reduced the extent of data augmentation during training, since data augmentation is proven effective in mitigating overfitting. For fair comparison, the training details are otherwise the same as in the supervised learning case, including learning rate and scheduling.

\subsection{Computing DSMI with input}

We compute $I_D(Z; X)$ the same way as we compute $I_D(Z; Y)$. By a simple change of variables, Eqn~\ref{eqn:DSMI} can be rewritten as:
\begin{equation*}
I_D(Z; X) = S_D(Z) - S_D(Z | X) = S_D(\mathbf{P}_Z, t) - \sum_{x_i \in X} p(X = x_i) S_D(\mathbf{P}_Z | X=x_i, t)
\end{equation*}

Compared to DSMI between the neural representation and the output, the DSMI with input is slightly more complicated because the input signals do not fall into discrete categories, i.e. $x_i$ are not naturally defined. To this end, for the set of $n$ input images, we flatten them and perform spectral clustering. For a fair comparison with $I_D(Z; Y)$, we cluster these vectors into the same number of categories as the number of classes in $Y$. The remaining process is the same as how we compute DSMI with output.

\subsection{Details on Imagenet cross-model correlation}
\subsubsection{Pre-trained models}
The pre-trained models are obtained from the Huggingface PyTorch Image Models repository~\citeSupp{timm}. Specifically, we used the models listed in \url{https://github.com/huggingface/pytorch-image-models/blob/main/results/results-imagenet.csv}, which the owners shared under Apache License 2.0. There are originally 1002 models, but 40 models failed our standardized pipeline, likely due to the presence of auxiliary structures, leading to a final set of 962 models. All of these models are pre-trained on ImageNet~\citeSupp{ImageNet}.

\subsubsection{ImageNet accuracy}
The ImageNet validation accuracy (top-1 and top-5) is taken from \url{https://github.com/huggingface/pytorch-image-models/blob/main/results/results-imagenet.csv}. Test accuracy (top-1 and top-5) is taken from \url{https://github.com/huggingface/pytorch-image-models/blob/main/results/results-imagenet-real.csv}.

\subsubsection{DSE and DSMI}
The full ImageNet validation set contains 1000 classes each with 50 samples. For efficient computation, we measured DSE $S_D(Z)$, DSMI $I_D(Z; X)$, and $I_D(Z; Y)$ on the union of two commonly used subsets of ImageNet validation set: namely Imagenette and Imagewoof. The former contains 10 very distinct classes (tench, English springer, cassette player, chain saw, church, French horn, garbage truck, gas pump, golf ball, parachute), while the latter contains 10 very similar classes (Australian terrier, Border terrier, Samoyed, Beagle, Shih-Tzu, English foxhound, Rhodesian ridgeback, Dingo, Golden retriever, Old English sheepdog). We believe that the union of these two subsets forms a representative subset of ImageNet.

\subsection{Logarithm base}
It can be verified that the choice of a logarithm base does not affect the validity of our formulations. By convention, we choose base 2 ($\log_2$) in our implementation. Hence, the unit of entropy will be bits.

\clearpage
\newpage
\section{All Raw Results}
\label{supp:all_raw_results}
Here we show the raw results corresponding to the following figures: Figures~\ref{fig:main_results_DSE}, \ref{fig:main_results_DSMI_output}, \ref{fig:main_results_DSMI_input}, \ref{fig:main_results_CSE}, \ref{fig:main_results_CSMI_output}, \ref{fig:main_results_CSMI_input}.

\begin{figure*}[!thb]
    \centering
    \includegraphics[width=\textwidth]{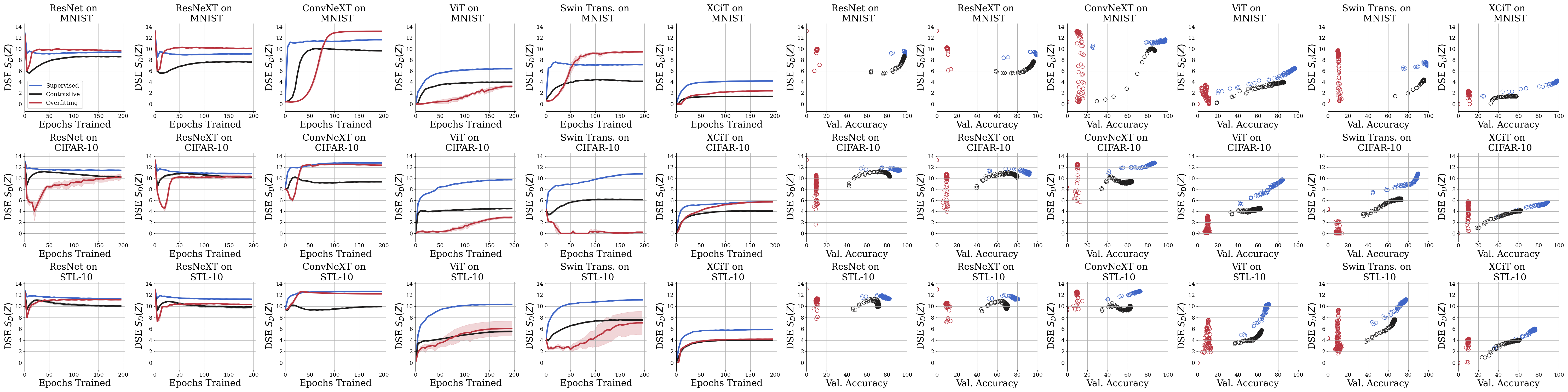}
    \caption{Raw results for Figure~\ref{fig:main_results_DSE}.}
    \label{fig:raw_results_DSE}
\end{figure*}

\begin{figure*}[!thb]
    \centering
    \includegraphics[width=\textwidth]{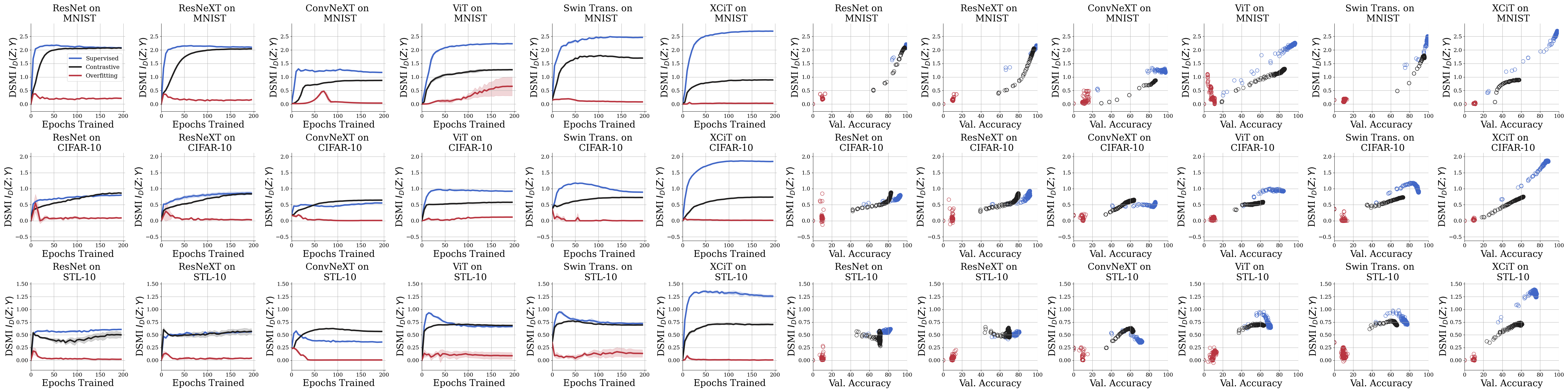}
    \caption{Raw results for Figure~\ref{fig:main_results_DSMI_output}.}
    \label{fig:raw_results_DSMI_Y}
\end{figure*}

\begin{figure*}[!thb]
    \centering
    \includegraphics[width=\textwidth]{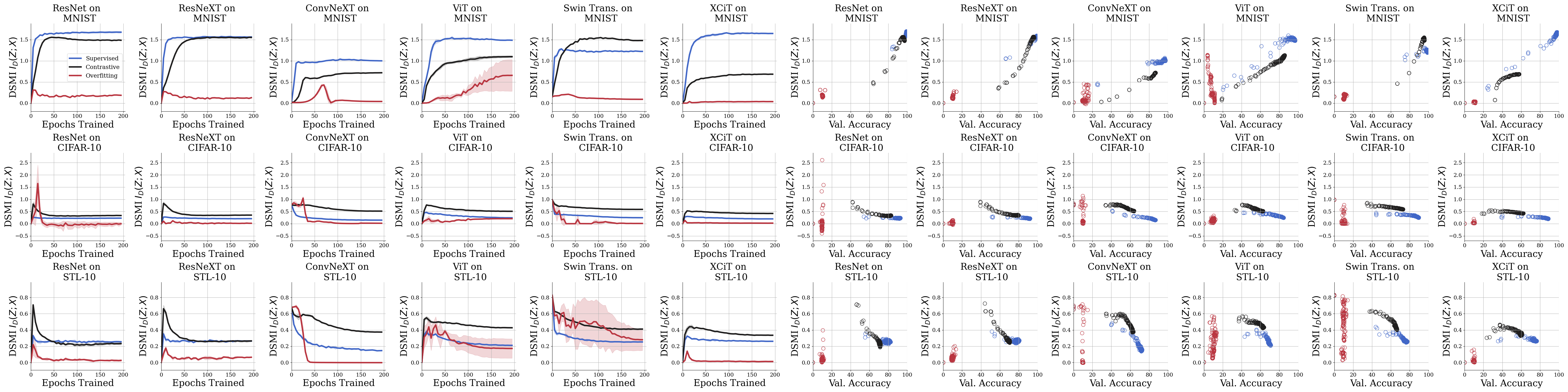}
    \caption{Raw results for Figure~\ref{fig:main_results_DSMI_input}.}
    \label{fig:raw_results_DSMI_X}
\end{figure*}

\begin{figure*}[!thb]
    \centering
    \includegraphics[width=\textwidth]{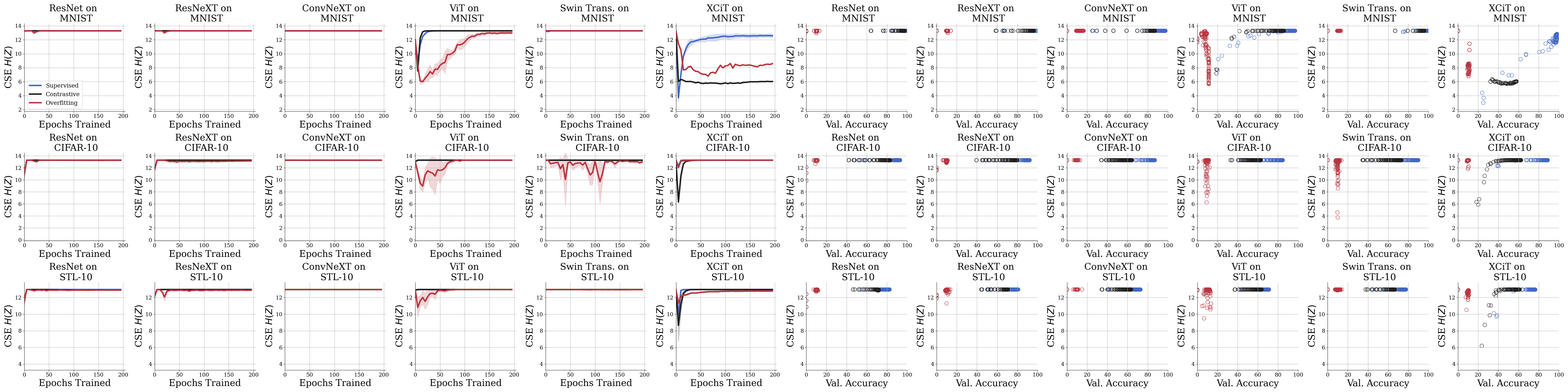}
    \caption{Raw results for Figure~\ref{fig:main_results_CSE}.}
    \label{fig:raw_results_CSE}
\end{figure*}

\begin{figure*}[!thb]
    \centering
    \includegraphics[width=\textwidth]{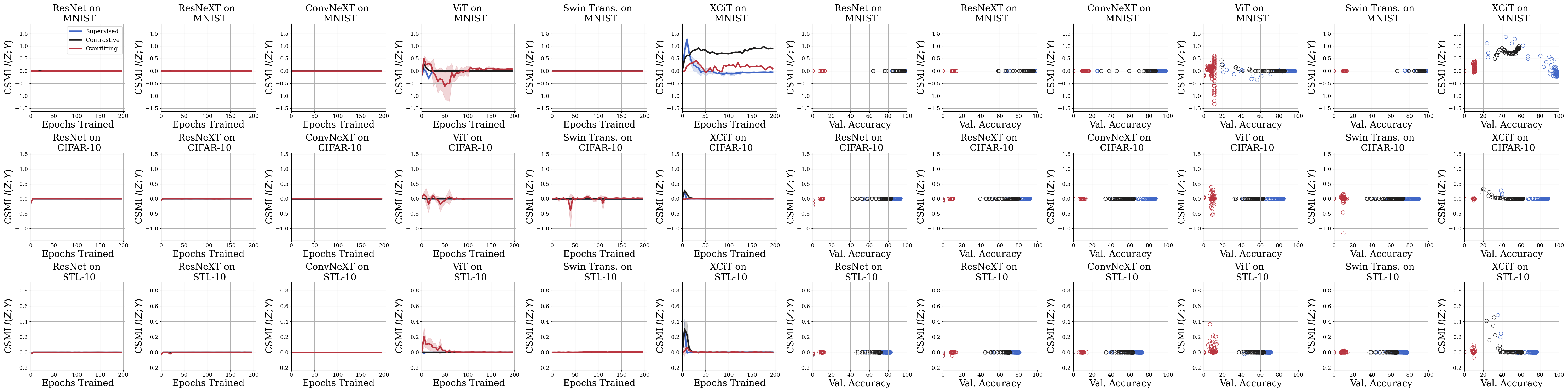}
    \caption{Raw results for Figure~\ref{fig:main_results_CSMI_output}.}
    \label{fig:raw_results_CSMI_Y}
\end{figure*}

\begin{figure*}[!thb]
    \centering
    \includegraphics[width=\textwidth]{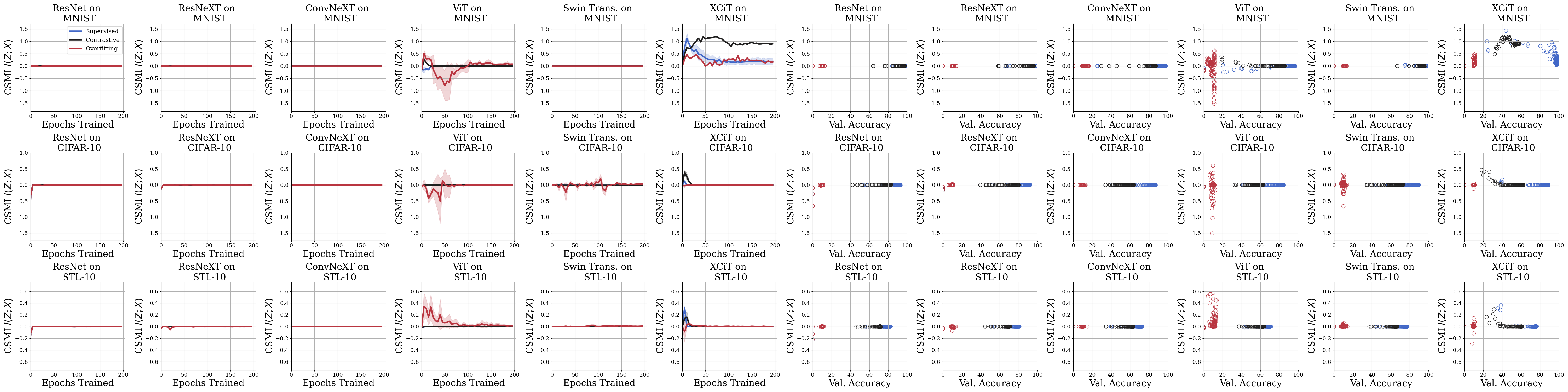}
    \caption{Raw results for Figure~\ref{fig:main_results_CSMI_input}.}
    \label{fig:raw_results_CSMI_X}
\end{figure*}

\clearpage
\newpage
\section{DSE Initialization Experiments}
\label{supp:dse_initialization}
For all experiments on the evolution of DSE and DSMI during neural network training, we initialized the model parameters using the same standard initialization process.

For convolutional layers (i.e. Conv and ConvTranspose), we used Kaiming normal initialization~\citeSupp{KaimingInit}. For linear layers (a.k.a. fully-connected layers), we initialized them with a normal distribution with mean 0 and standard deviation 1e-3. For batch normalization weights, we initialized them with a constant value of 1.0. For all biases, we initialized them with a constant value of 0.

Despite the fact that we used the same initialization, the initial value of DSE is not always the same across different experiments. For example, in Figure~\ref{fig:raw_results_DSE}, ResNet on MNIST began at a high DSE, while ViT on MNIST began at a low DSE. While this does not change the general increasing trend of DSE during training, we wonder: \textbf{Will initializing the network at a high DSE versus a low DSE affect the learning process?}

\begin{figure}[!thb]
  \begin{minipage}[c]{0.32\textwidth}
    \includegraphics[width=\textwidth]{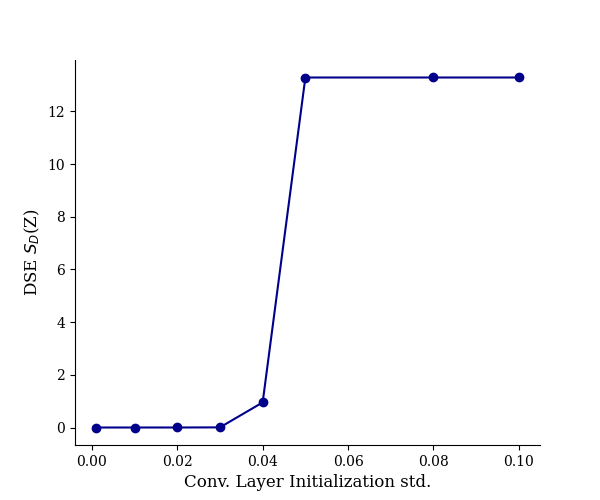}
    \caption{Initialization parameters affects the initial DSE (CIFAR-10).}
    \label{fig:DSE_init_stds_cifar10_resnet}
  \end{minipage}\hfill
  \begin{minipage}[c]{0.66\textwidth}
    \includegraphics[width=\textwidth, height=0.4\textwidth]{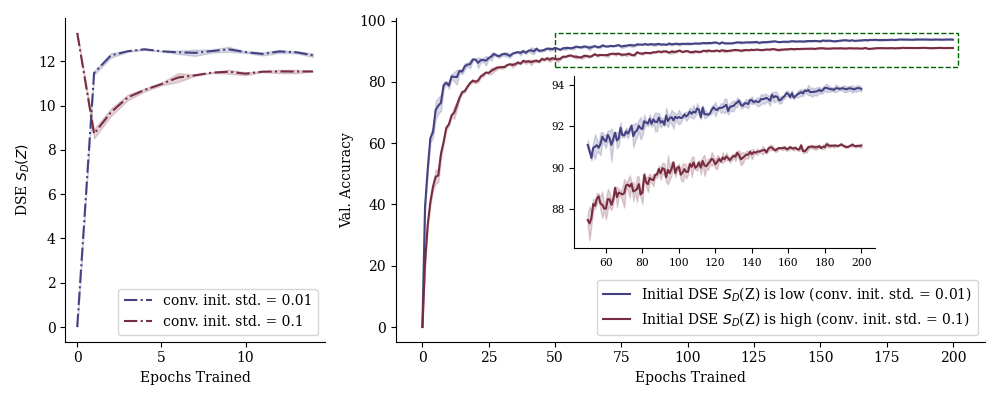}
    \caption{Initializing the network at a low DSE allows for faster convergence and better final performance (CIFAR-10).}
    \label{fig:DSE_init_compare_cifar10_resnet_supp}
  \end{minipage}
\end{figure}

After further investigation, we found that we can control this behavior by modifying the initialization parameters on the convolutional layers. Specifically, we replaced the initialization method for the convolutional layers with a normal distribution with a mean of 0 and a tunable standard deviation. We observed a monotonic trend in the initial DSE as we adjusted the standard deviation (Figure~\ref{fig:DSE_init_stds_cifar10_resnet}). However, as the transition is abrupt, it is not easy to target an arbitrary DSE value.

\begin{figure}[!thb]
  \begin{minipage}[c]{0.32\textwidth}
    \includegraphics[width=\textwidth]{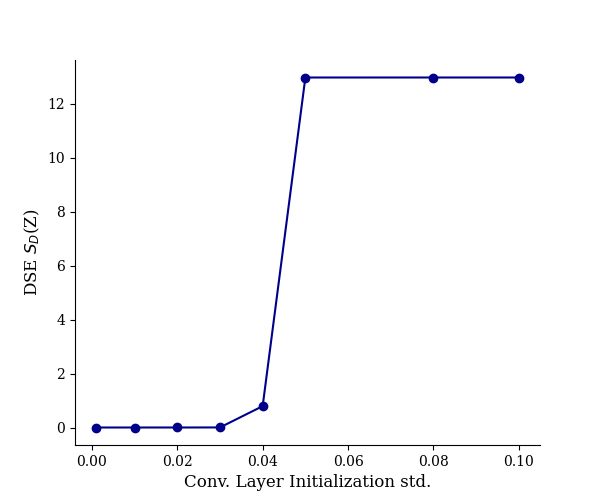}
    \caption{Initialization parameters affects the initial DSE (STL-10).}
    \label{fig:DSE_init_stds_stl10_resnet}
  \end{minipage}\hfill
  \begin{minipage}[c]{0.66\textwidth}
    \includegraphics[width=\textwidth, height=0.4\textwidth]{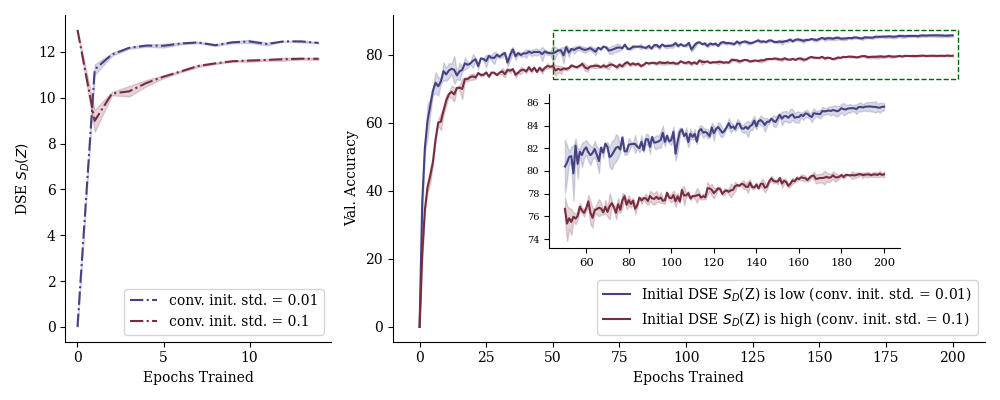}
    \caption{Initializing the network at a low DSE allows for faster convergence and better final performance (STL-10).}
    \label{fig:DSE_init_compare_stl10_resnet_supp}
  \end{minipage}
\end{figure}

To answer the \textbf{question in bold} above, we trained two ResNets on the CIFAR-10 dataset, one starting with a low DSE~(conv. init. std. = 0.01) and the other starting with a high DSE~(conv. init. std. = 0.1). Experiments are repeated under 3 random seeds. Intuitively, we would expect the former to converge faster than the latter, because in the former case the embedding manifold only needs to monotonically increase DSE~(Figure~\ref{fig:DSE_init_compare_cifar10_resnet_supp} left, blue line), while in the latter case the DSE needs to first decrease from the peak and then increase~(Figure~\ref{fig:DSE_init_compare_cifar10_resnet_supp} left, red line). Indeed, this intuition is supported by empirical results. In the right panel of Figure~\ref{fig:DSE_init_compare_cifar10_resnet_supp}, it can be seen that, at least in our preliminary experiments, \textbf{initializing the network at a low DSE allows faster convergence and better final performance.}

We conducted the same experiments on STL-10 and observed the same results. The corresponding figures are Figure~\ref{fig:DSE_init_stds_stl10_resnet} and Figure~\ref{fig:DSE_init_compare_stl10_resnet_supp}.

\begin{figure}[!thb]
    \centering
    \includegraphics[width=\textwidth]{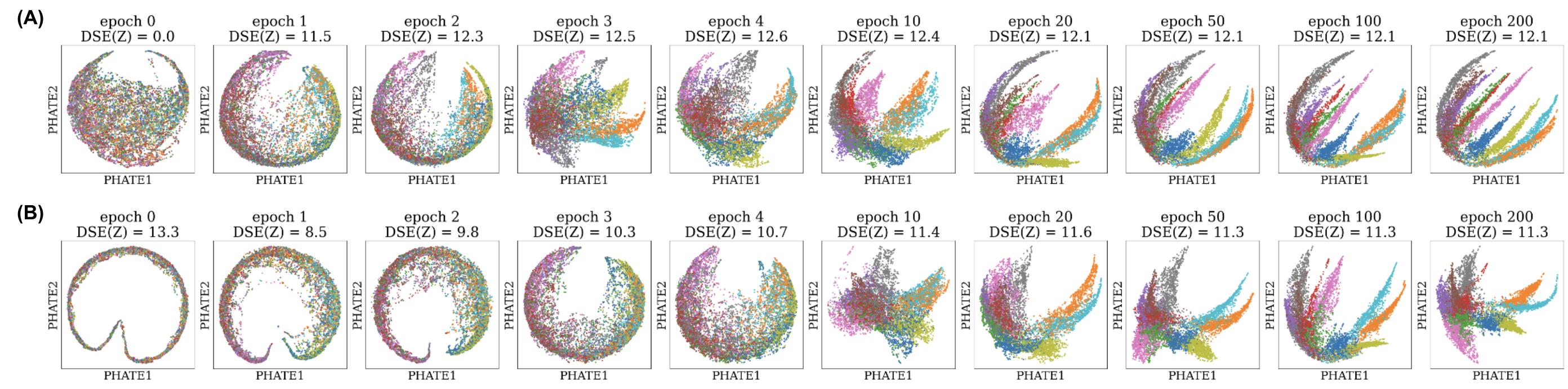}
    \caption{PHATE representation of the embedding spaces during training for low (panel A) and high (panel B) initial DSE. Colors represent ground truth class labels.}
    \label{fig:DSE_init_phate_cifar10_resnet}
\end{figure}

This observation can be further corroborated by the visualizations of the embedding space. When we initialize the network at a low DSE~(Figure~\ref{fig:DSE_init_phate_cifar10_resnet}(A)), the PHATE~\citeSupp{phate} representation of the embedding space is close to a uniform blob. As the network undergoes more training, the embedding space becomes organized into distinct branches, each corresponding to a different class. This process is relatively smooth and monotonic.

On the other hand, when we initialize the network at a high DSE~(Figure~\ref{fig:DSE_init_phate_cifar10_resnet}(B)), the PHATE representation almost looks like a curved ribbon. As training progresses, the embedding space first expanded to become more blob-like and then stretches out into branches. Compared to the other case above, which only has a branching phase, this process has an additional expansion phase before the branching phase, which may explain the slower convergence. Moreover, the final representations are not as clean and well-separated as the other case.

All of these results together suggest that different network initializations may affect the convergence rate and final performance, and that some properties of network initialization can be captured by DSE. However, since this is not the main focus of this paper, we leave it to future research to explore this topic in more depth.

\vskip 0.5in
\section{Additional DSE Results: Intuition and MAGIC visualization}
\label{supp:DSE_additional_results}
\subsection{DSE vs. Diffusion $t$ in the Intuition Figure}
\begin{figure*}[!htb]
    \centering
    \includegraphics[width=\textwidth]{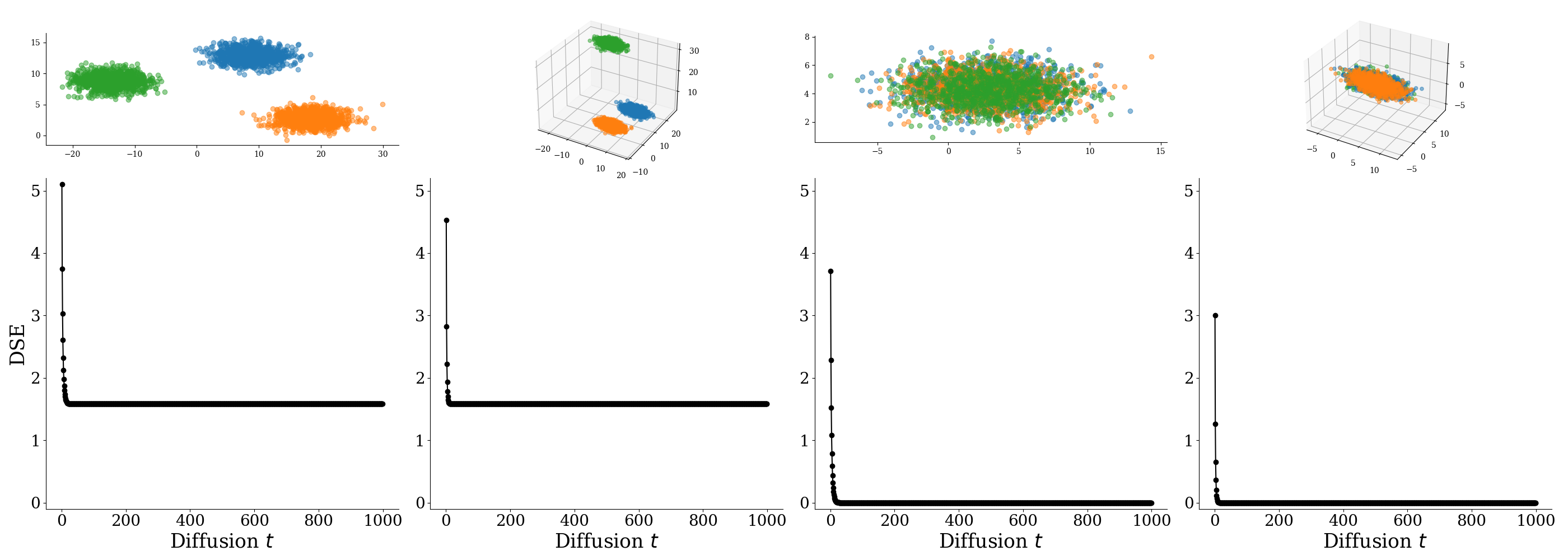}
    \caption{\textbf{DSE vs. Diffusion $t$ for Figure 1.} \textbf{Left:} $X$ are $k$ separated clusters. Simulated with 3 blobs in 2 or 3 dimensions. $S_D(\mathbf{P}_X,t) \approx \log(k) \text{ when t increases}$ \textbf{Right:} $X$ is a single blob. Simulated with 1 blob in 2 or 3 dimensions. $S_D(\mathbf{P}_X,t) \approx 0 \text{ with increasing t}$.}
    \label{fig:DSE_intuition_simulation}
\end{figure*}

Figure~\ref{fig:DSE_intuition_simulation} illustrates DSE vs. diffusion time $t$ for the cases in the intuition figure (Figure~\ref{fig:intuition}). It can be seen that DSE approaches $log_2(3) \approx 1.585$ with 3 nicely separated blobs, while it approaches zero with one single blob.

\subsection{MAGIC Visualization for DSE vs. Diffusion $t$}
\begin{figure*}[!htb]
    \centering
    \includegraphics[width=\textwidth]{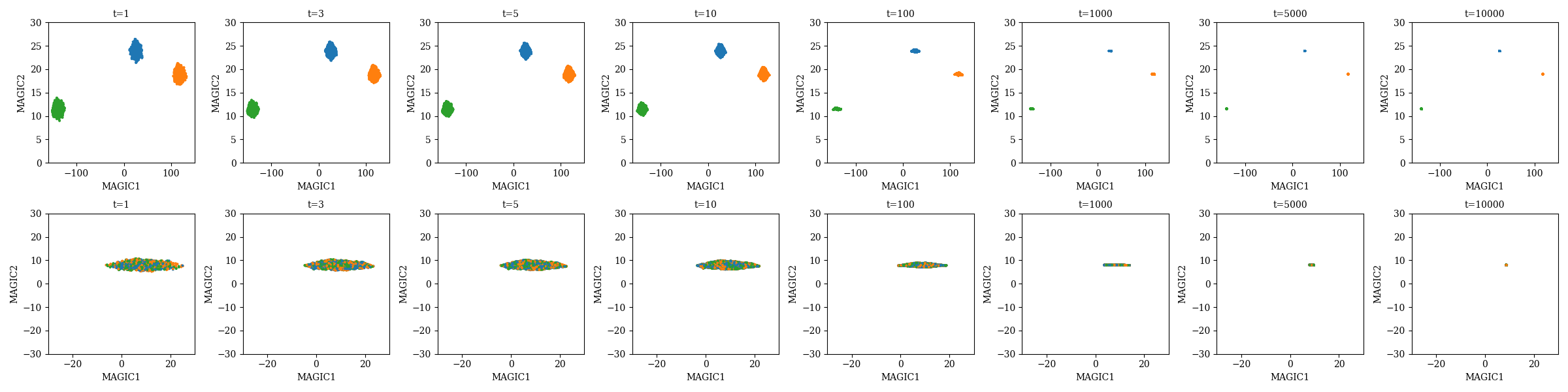}
    \caption{\textbf{MAGIC representation of simulated blobs data in 3 dimension.} \textbf{Top panel:} $X$ are $k$ separated clusters. \textbf{Bottom panel:} $X$ is a single blob. As diffusion $t$ increases, a dataset characterized by three distinct blobs evolves into three separate ``lines." Conversely, a dataset featuring a singular blob transforms into a single ``line". }
    \label{fig:visualize_DSE_intuition}
\end{figure*}

To emphasize that DSE reflects the structure of the data manifold instead of data variations in the ambient space, we employed MAGIC \cite{magic} for visual representation. This visualization technique showcases the diffusion geometric representation of the data manifold across various diffusion $t$ values. As evident in Figure~\ref{fig:visualize_DSE_intuition}, with an increase in diffusion $t$, a dataset characterized by three distinct blobs evolves into three separate `lines'. Conversely, a dataset featuring a singular blob transforms into a single "line." This visual representation using MAGIC underscores the inherent structures within the data manifold and it is evident that DSE aligns well with these visual interpretations.

\subsection{Ablation Study}
We conducted an ablation study on diffusion spectral entropy by experimenting with alternative methods for computing entropy. In addition to diffusion spectral entropy (DSE) and classic Shannon entropy (CSE), we investigate three alternatives: diffusion matrix entry entropy (DMEE) in which the classic Shannon entropy is computed on the entries in the diffusion matrix $\mathbf{P}$; entropy with eigenvalues from the binary adjacency matrix of k-nearest-neighbor graph; entropy with eigenvalues from the weighted adjacency matrix $\mathcal{G}$ constructed with the Gaussian kernel.

\begin{figure*}[!htb]
    \centering
    \includegraphics[width=0.8\textwidth]{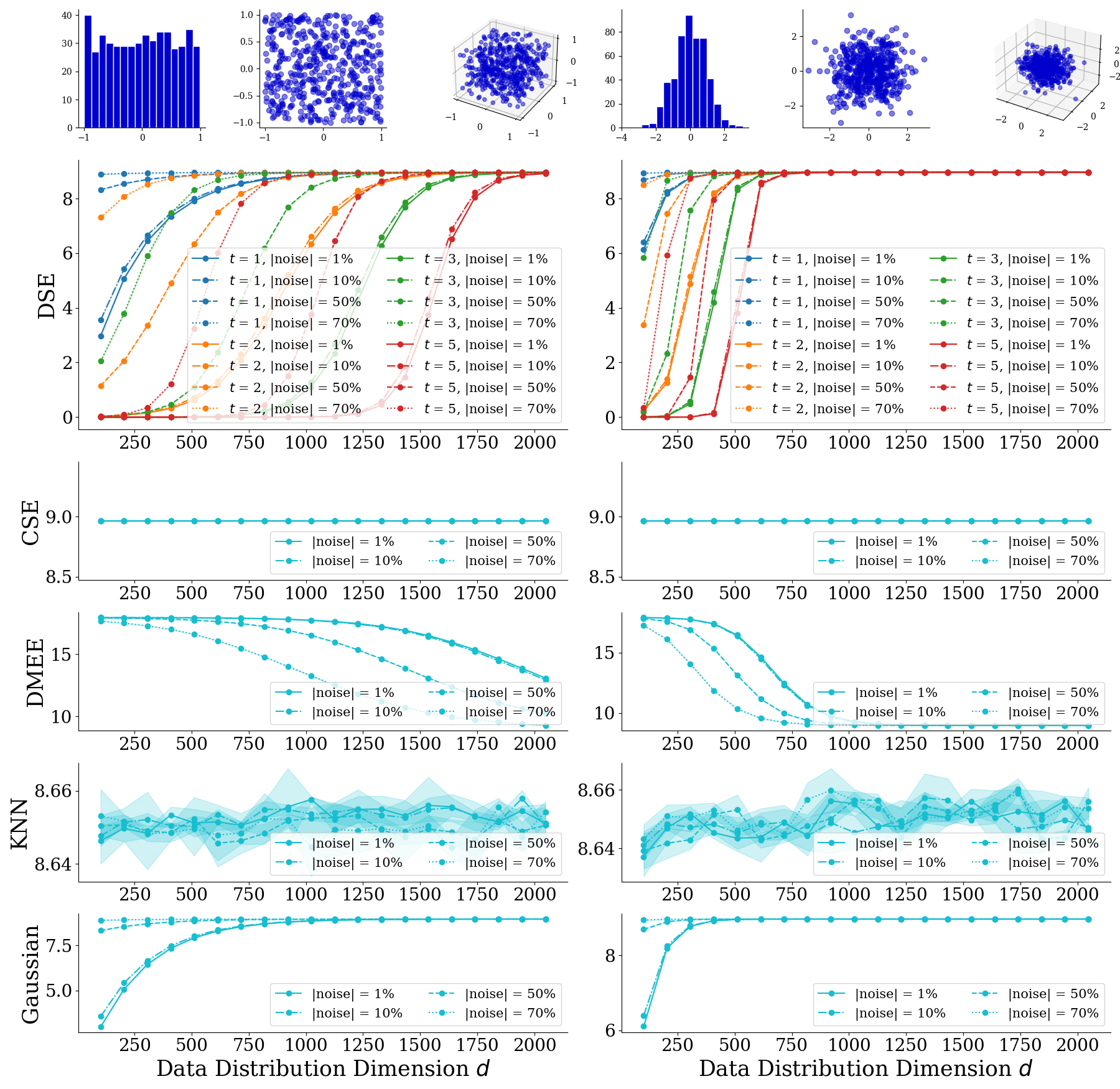}
    \caption{\textbf{Ablation study on diffusion spectral entropy (DSE)}.}
    \label{fig:DSE_ablation_study}
\end{figure*}

\begin{table*}[!htb]
    \caption{Descriptions of methods being compared.}
    \centering
    \begin{tabular}{cc}
    \toprule
    Method & Description \\
    \midrule
    DSE & diffusion spectral entropy: proposed method (Eqn~\ref{eqn:diffspectralentropy}) \\
    CSE & classic Shannon entropy (Eqn~\ref{eqn:shannon_entropy})\\
    DMEE & diffusion matrix entry entropy (Shannon entropy over $\{\mathbf{P}_{i,j}\}$) \\
    KNN & with eigenvalues from binary adjacency matrix of k-nearest-neighbor graph\\
    Gaussian & with eigenvalues from the weighted adjacency matrix $\mathcal{G}$, where $\mathcal{G}(z_1, z_2) = e^{-\frac{\|z_1 - z_2\|^2}{\sigma}}$\\
    \bottomrule
    \end{tabular}
\label{tab:full}
\end{table*}

We visualized the behavior of different entropy methods against the intrinsic dimension of the data. We can see from Figure \ref{fig:DSE_ablation_study} that diffusion matrix entry entropy (DMEE) decreases as the intrinsic dimension of the data increases. The entropy computed on eigenvalues from the binary adjacency matrix of the k-nearest-neighbor graph (KNN method) jitters around a fixed value as the intrinsic data dimension increases. The entropy on the eigenvalues of the weighted adjacency matrix $\mathcal{G}$ starts with a high entropy in a relatively small data dimension and continues to increase until it becomes stagnant. It plateaus in much smaller dimensions on both uniform and Gaussian distributed data. On data with a high percentage of noise (50\%, 70\%), we observe that entropy on eigenvalues of Gaussian adjacency matrix barely changes despite increasing data dimension, while diffusion spectral entropy (DSE) can remain resilient to noise and effectively reflect the increasing intrinsic dimension as the diffusion parameter $t$ is adjusted.

\vskip 0.5in
\section{Additional DSMI Results: Intuition and Subsampling Robustness}
\label{supp:DSMI_additional_results}

\subsection{DSMI vs. Diffusion $t$ in the Intuition Figure}
\begin{figure*}[!htb]
    \centering
    \includegraphics[width=\textwidth]{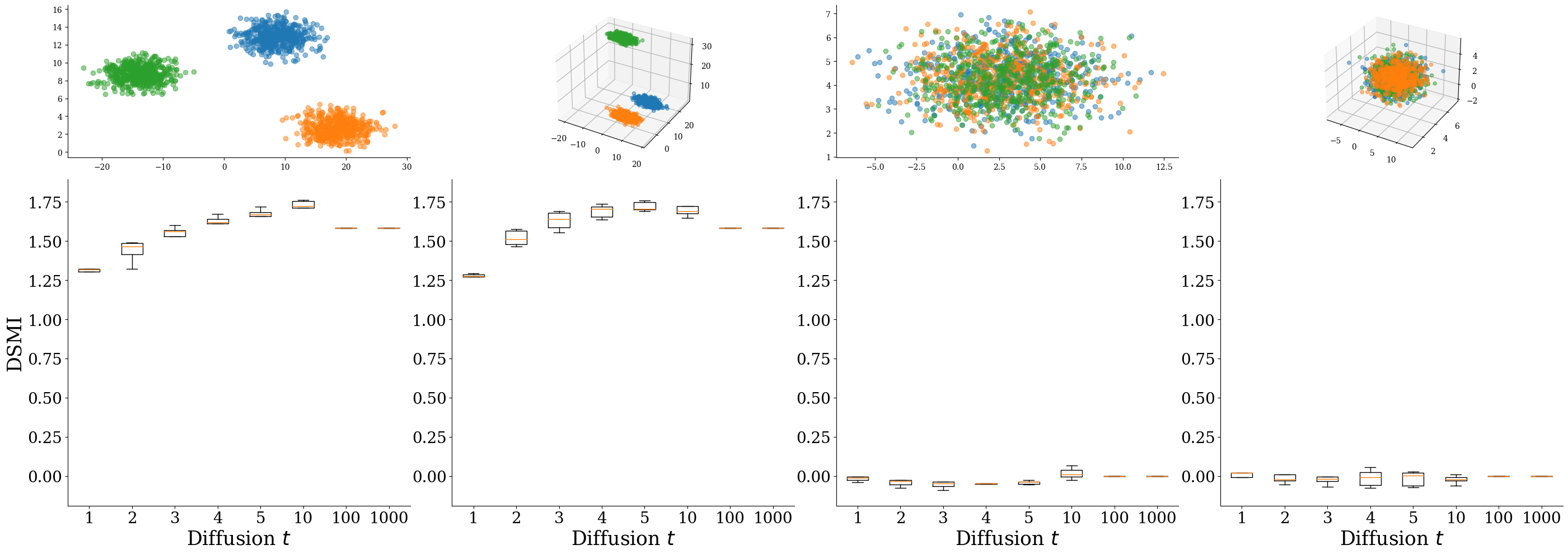}
    \caption{\textbf{DSMI simulations for Figure~\ref{fig:intuition}. }\textbf{Left:} $Y$ is on separated clusters. Simulated with 3 blobs in 2 or 3 dimensions. $S_D(X|Y) < S_D(X)$ and $I_D(X; Y)$ are positive. \textbf{Right:} $Y$ is close to a uniform sub-sampling of $X$. Simulated with 1 blob in 2 or 3 dimensions. $I_D(X; Y)$ is around 0, but can have negative values if $S_D(X|Y)$ is larger than $S_D(X)$ for numeric reasons.}
    \label{fig:DSMI_intuition_simulation}
\end{figure*}

Figure~\ref{fig:DSMI_intuition_simulation} illustrates DSMI for the cases in the intuition figure (Figure~\ref{fig:intuition}). It can be seen that DSMI is positive on nicely separated clusters, while close to zero on well-mixed clusters. Moreover, when $t$ is sufficiently high, DSMI approaches $\log(k)$, which in this case is $log_2(3) \approx 1.585$.

\subsection{Subsampling Technique for DSMI: Subsampling Robustness}

As mentioned in \textbf{Definition~\ref{def:DSMI}}, we use a subsampling technique to compute \textit{Diffusion Spectral Mutual Information~(DSMI)}.

\begin{figure*}[!htb]
    \centering
    \includegraphics[width=0.6\textwidth]{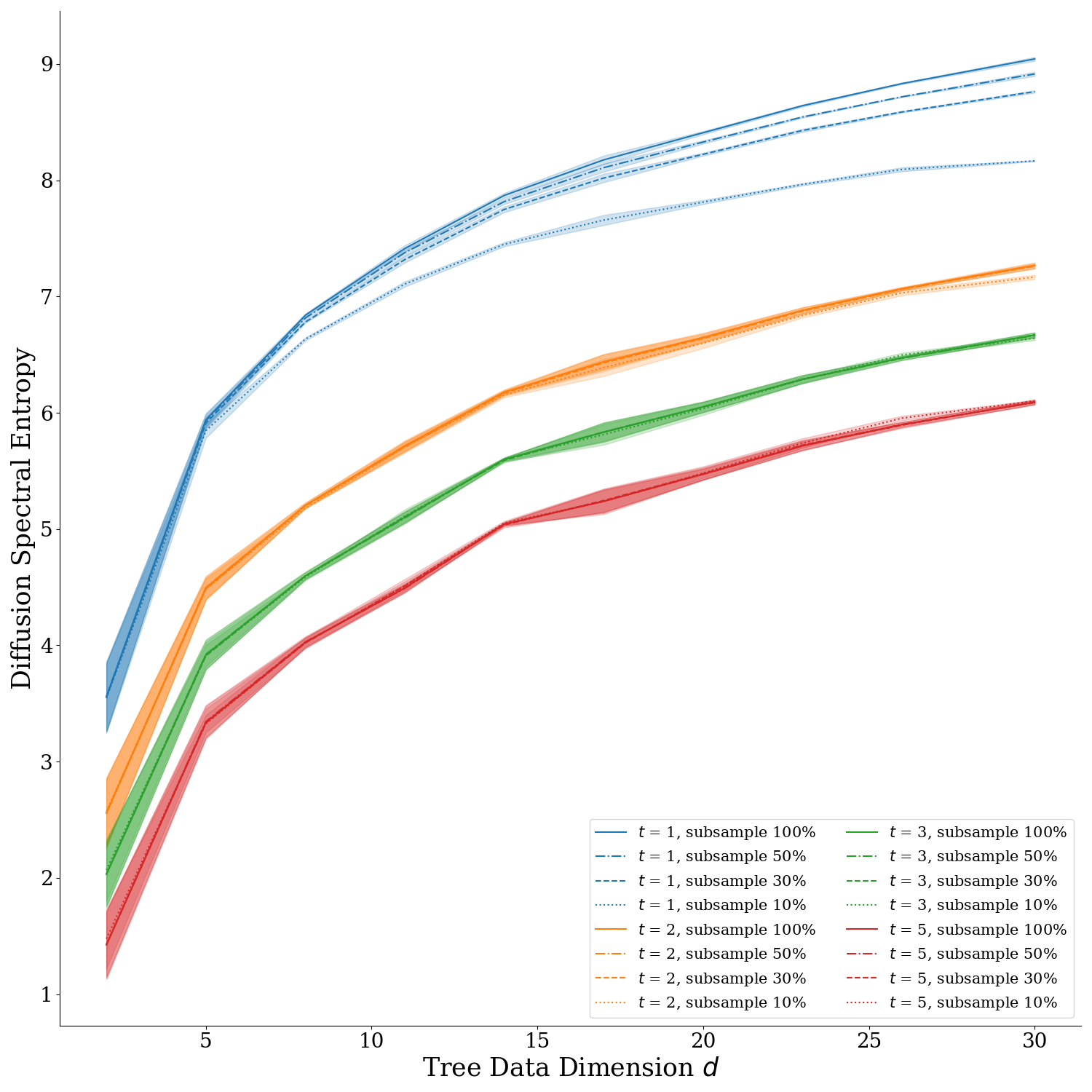}
    \caption{\textbf{Diffusion spectral entropy estimation is robust to subsampling.} Subsampled $S_D(Z)$ are computed on synthetic multi-dimensional trees.}
    \label{fig:DSE_subsampling}
\end{figure*}

Since DSMI is computed as a conditional DSE, we hereby justify this technique by showing that Diffusion Spectral Entropy~(DSE) is robust to subsampling (Figure~\ref{fig:DSE_subsampling}). Four line styles of the same color, respectively, represent different subsampling ratio ranging from 100\% (full dataset) to 10\%. It can be observed that DSE can be reliably computed despite significant subsampling.

On the other hand, we also need to point out that the subsampling robustness is constrained by the DSE upper bound. DSE is capped by $\log_2 n$ where $n$ is the number of data points. For example, if the entire population has a DSE of 8 while we subsample fewer than $2^8 = 256$ data points, the subsampled DSE will be an underestimate. In our studies, the subsampling ratio is about 10\%, which always yields a significant number of data points per category. This technique shall be used with caution if it is applied to other studies with more extreme subsampling.

\clearpage
\newpage

\clearpage
\newpage

\bibliographySupp{references}
\bibliographystyleSupp{apalike}

\vspace{12pt}

\end{document}